\documentclass[sigconf, letterpaper]{acmart}

\usepackage{libertine}
\usepackage{threeparttable}
\usepackage{multirow}
\usepackage{stmaryrd}
\usepackage{graphicx}
\usepackage{amsmath}
\usepackage[linesnumbered, algoruled,longend,lined]{algorithm2e}
\usepackage{array}
\usepackage{url}
\usepackage{color}

\DeclareMathOperator*{\argmax}{argmax}

\begin{document}
\title[Power Side-Channel Attack on Convolutional Neural Network Accelerators]{I Know What You See: Power Side-Channel Attack on Convolutional Neural Network Accelerators}
\author{Lingxiao Wei$^{\dag}$, Bo Luo$^{\dag}$, Yu Li$^{\dag}$, Yannan Liu$^{\dag\ddag}$ and Qiang Xu$^{\dag}$}
\affiliation{
  \institution{$^\dag$\underline{CU}hk \underline{RE}liable Computing Laboratory (CURE) \\
  Department of Computer Science and Engineering \\
  The Chinese University of Hong Kong, Shatin, N.T., Hong Kong \\
  $^\ddag$Sangfor Technologies Inc., Shenzhen, China \\
  Email: \{lxwei, boluo, yuli, ynliu, qxu\}@cse.cuhk.edu.hk}
}






\copyrightyear{2018}
\acmYear{2018}
\setcopyright{acmcopyright}
\acmConference[ACSAC '18]{2018 Annual Computer Security Applications Conference}{December 3--7, 2018}{San Juan, PR, USA} 
\acmBooktitle{2018 Annual Computer Security Applications Conference (ACSAC '18), December 3--7, 2018, San Juan, PR, USA}
\acmPrice{15.00} 
\acmDOI{10.1145/3274694.3274696} 
\acmISBN{978-1-4503-6569-7/18/12}

\renewcommand{\shortauthors}{L. Wei et al.}



\begin{abstract}
Deep learning has become the \emph{de-facto} computational paradigm for various kinds of perception problems, including many
privacy-sensitive applications such as online medical image analysis. No doubt to say, the data privacy of these deep learning systems is a serious concern.
Different from previous research focusing on exploiting privacy leakage from deep learning models, 
in this paper, we present the first attack on the implementation of deep learning models. 
To be specific, we perform the attack on an FPGA-based convolutional neural network accelerator and we manage to recover the input image from the collected power traces without knowing the detailed parameters in the neural network.
For the MNIST dataset, our power side-channel attack is able to achieve up to 89\% recognition accuracy.
\end{abstract}

%
%

  

\maketitle

\section{Introduction}\label{sec:intro}

Deep neural network (DNN) is widely used in many safety-critical and security-sensitive artificial intelligence (AI) applications such as biometric authentication, autonomous driving, and financial fraud analysis. Consequently, their security is a serious concern and requires urgent research attention.

Recent research has shown the security and privacy of DNN models can be compromised. 
In~\cite{PapernotMJFCS16}, attackers deceive the DNN system to make misclassifications by adding small perturbation to the original images.
In~\cite{FredriksonJR15}, malicious parties are able to recover private images that are used to train a face recognition system by  analyzing the outputs of the DNN model.
However, the success of these attacks requires the full knowledge of the DNN model, which is not usually available in real-world applications as model parameters are  valuable assets for deep learning tasks and are always kept confidential. 
In addition, a variety of privacy-leaking prevention techniques~\cite{AbadiCGMMT016, MohasselZ17, PapernotM0JS16} has emerged to mitigate these attacks.

\begin{figure}[thb]
\centering
\includegraphics[width=1.0\linewidth]{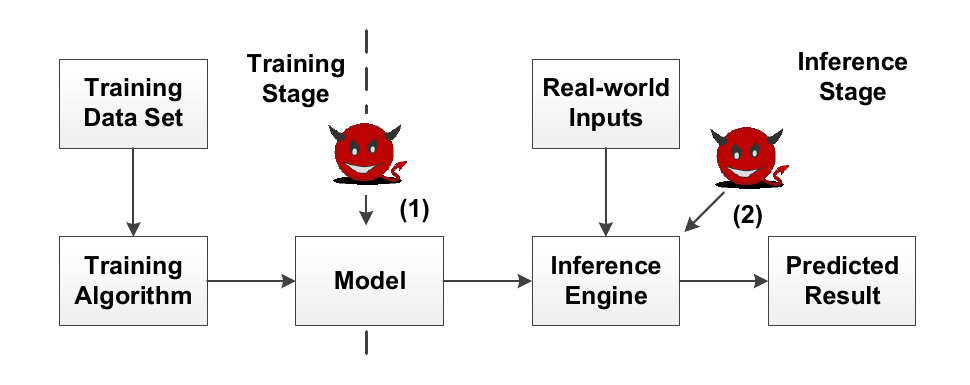}
\caption{Deep learning flow: at training stage, a training algorithm generates a prediction model from a large number of training examples, while at inference stage, a inference engine predict the result from real-world inputs with the trained model. (1) existing research assumes attacks on DNN models (2) our approach attacks the hardware of inference engine.}
\label{fig:ml_threat}
\end{figure}

As illustrated in Fig.~\ref{fig:ml_threat},
apart from the privacy leakage from DNN models, the information of real-world inputs can also be leaked at inference stage which remains largely unexplored before. 
In privacy-sensitive systems, the direct access to the input data of the inference engine is often strictly controlled so that it is nearly impossible for potential adversaries to retrieve private information in a stealthy manner. 
For instance, encrypted medical images are provided to the DNN inference engine and they are decrypted inside the inference engine to prevent eavesdropping on the communication paths.
Under such circumstances, we are concerned whether attackers can use side channel information (e.g., power consumption) to retrieve private data.
Dedicated DNN accelerators are expected to gain mainstream adoption in the foreseeable future due to their high computation efficiency~\cite{sze2017efficient}.
In this paper, we present a power side-channel attack on an FPGA-based convolutional neural network (CNN) accelerator which performs the task of image classification.
The attack target on the hardware component executing the \emph{convolution operation in the first layer of CNN} which is usually implemented by \emph{line buffer}, a common structure in many image-related processing hardware.
The primary objective of our attack is to recover the private input image from eavesdropping the power consumption of the neural accelerator when it performs calculations for the first layer.

To the best of our knowledge, the proposed attack is the first one that exploits the privacy leakage in neural accelerators using power side channels. Particularly, unlike previous privacy attacks targeting at reproducing samples in the training set using model outputs~\cite{FredriksonJR15}, our proposed attack aims to recover the input being inferenced and we do not assume the pre-requisite knowledge of model parameters or model outputs. The main contributions of our work include: 
\begin{itemize}
\item Power side channel is often quite noisy, and the collected power trace contains distortions brought by various circuit components. We present a novel power extraction technique to precisely recover the power consumption for each clock cycle. 
\vspace{3pt}
\item We develop novel algorithms to retrieve each pixel value of the input image.  To be specific, as the convolution operation only relates to a limited number of pixels, we develop algorithms to infer the values of pixels either from power traces directly or from a pre-built power-pixel template. Finally, the image can be reconstructed by piecing all inferred pixels together. 
\end{itemize}

The remainder of this paper is organized as follows: Section~\ref{sec:background} introduces the background knowledge with threat model follows in Section~\ref{sec:threat_model}.
Next, we give an overview on the proposed attack flow in Section~\ref{sec:overview} for two attack scenarios. We introduce how to accurately estimate the power from the noisy power side channels in Section~\ref{sec:power_extract}. The details of the two attack scenarios are then introduced in Section~\ref{sec:background_detection} and
in Section~\ref{sec:power_templates}, respectively.
Finally, we discuss related work in Section~\ref{sec:related_work} and conclude this work in Section~\ref{sec:conclusion}. In addition, to illustrate the preliminaries of our proposed attack more clearly, we give a detailed introduction of the concept of convolutional neural network and the design of neural accelerators. Also, we discuss the power measurement setup and characterization of the neural accelerators in Section~\ref{sec:appendix-background} in the Appendix. Then we talk about the limitation and countermeasures in Appendix~\ref{sec:discussion} and give the results of our attack on the MNIST dataset in Appendix~\ref{sec:appendix-mnist}.

\section{Background}\label{sec:background}
In this section, we give a brief introduction to convolutional neural network, neural accelerators and the power characterization of the accelerator, respectively. For more information, readers can refer to Section~\ref{sec:appendix-background} in the Appendix.

\noindent\textbf{Convolutional Neural Network:} Convolutional neural network has been the prevalent network architecture in image-related processing tasks. It is constructed by a sequence of layers where the first few layers perform the \emph{convolutional operation}. In this attack, we focus on the first layer of the convolutional neural network whose input is the raw image and the computation in this layer is convolution. Generally speaking, convolution layer uses a 2-D filter (i.e., kernel) to slide over the 2-D input image (e.g., gray image) and finally produces another 2-D image (i.e., feature map) using the convolution operation. Readers can resort to Fig.~\ref{fig:bnn_fpga} (a) for an illustration of the convolution operation.

\noindent\textbf{Neural Accelerator:} The computation in the convolution neural network is usually implemented by \emph{neural accelerators} on devices with stringent power budget. They are often based on FPGA or ASIC, and there are many design available in both academia and industry~\cite{qiu2016going, conti2015ultra, zhang2017frequency}. Our target is the convolutional layer, whose functionality is usually implemented by the \emph{line buffer}. \emph{Line buffer} is composed of buffer lines and a computation unit: \emph{buffer lines} are used to cache the pixels in recent lines while the \emph{computation unit} calculates the convolution result given the pixels and filter values. In this paper, we follow the design proposed by Zhao et al~\cite{ZhaoSZXLSGZ17}. They implement an accelerator for a compressed version of CNN~\cite{HubaraCSEB16} on FPGA.

\noindent\textbf{Power characterization of neural accelerator: } To accurately estimate the power constitution of line buffer, we simulated our design with Xilinx XPower analyzer and discovered the power in the convolution unit occupied more than 80\% of the total consumption regardless of the configuration of line buffer. Reader may refer to the appendix for a complete comparison among different line buffer configurations. Therefore, we can regard the measured power as a coarse-grain estimate for the power of convolution unit.

\begin{figure*}
\centering
\includegraphics[width=1.0\linewidth]{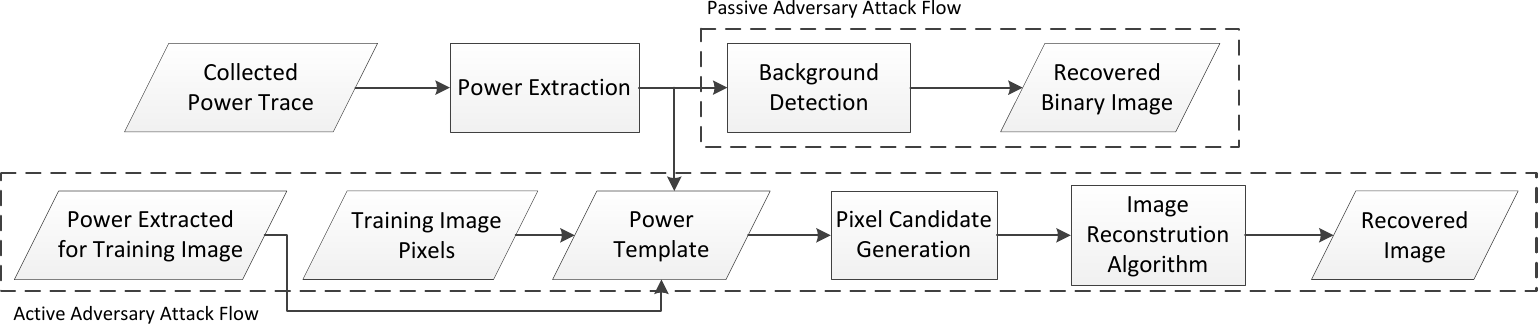}
\caption{Overview of the attack flow.}
\label{fig:overview}
\end{figure*}

\section{Threat Model}\label{sec:threat_model}
\noindent \textbf{Scenario:}
The primary goal of adversaries is to recover the input that fed into a convolutional neural accelerator which contains sensitive private information. Instead of reconstructing the samples in the training set, the adversaries try to reproduce the online input when the neural accelerator is actually performing the inference operation.
To facilitate the attack, we consider the adversaries come from the DNN accelerator design team or they are insiders in the companies hosting DNN infrastructures so that they are capable to monitor the power side channel.
At inference stage, DNN model designers usually deploy their trained model on a machine-learning service operator, such as BigML~\cite{BIGML} or Microsoft Azure~\cite{MSAZURE}, who uses dedicated DNN accelerators for the inference operation. 
They may also put their models on computing platforms (e.g., Qualcomm's Snapdragon 835~\cite{QSSNAP835}) with DNN-accelerating hardware. 
In many applications, 
for privacy concerns,
the inputs to the DNN accelerators are often protected with strict access control policies or strong encryption schemes. For instance, they may adopt secure processors, like AEGIS~\cite{SuhCGDD03}, to keep the incoming data confidential both in memory and on the disk. 
Thus, for attackers it is very difficult to obtain the inputs directly.
However, the side channels, especially the power side channel, are exposed unprotected to malicious insiders in the host of machine learning service company and DNN accelerator design team. 
They are capable to access to the power side channel output via malicious implanted Trojans or measurement circuits when the accelerator is actually running with real-world users' inputs.

\noindent
\textbf{Capability:} 
Firstly, we assume attackers are knowledgeable about the structure of the neural network and the input image size, but not the detailed parameters in the network. To be specific, for the targeted first convolution layer, the attackers need to know its filter size, number of input feature maps, and number of output feature maps. 
Secondly, adversaries can acquire the power trace of the DNN accelerator in high resolution either by oscilloscope measurement or power-monitoring Trojan.
We consider these two assumptions practical because firstly, many image-related tasks adopt existing neural network architecture (e.g., VGG-16/19 and ResNet) whose structure (including number of layers and configurations of each layer) is fixed and public and secondly, from the perspective of insiders, it is easy to implant Trojans or measurement circuits to get the power traces at runtime.
Thirdly, according to the ability of launching inference operation freely, we further divide the adversaries into two categories: \emph{passive adversary} and \emph{active adversary}.
Passive adversary can only eavesdrop on the power consumption when an input is processed by the DNN accelerator at inference stage.
Active adversary has an extra capability of profiling the relationship between power and input pixels by freely launching inference operation with arbitrary inputs on the targeted accelerator. The profiling phase can only be carried out prior to any actual calculation of user's private data. The main difference between passive and active attackers is the time that they get access to the power channel of the accelerator. We regard both attackers are realistic as only if they are physically accessible to the accelerator.

\section{Overview}\label{sec:overview}
The primary goal in this paper is to recover the input image from power traces of the targeted CNN accelerator.
The reason we choose the convolution in the first layer as attack target is as follows: firstly it 
directly processes the input image so the power obtained closely relates to the input.
Secondly, the inherent characteristic of convolution, which performs computation on a small bunch of pixels, can reduce the effort needed to infer the pixel values.

To evaluate the proposed attack, we implement a CNN accelerator~\cite{ZhaoSZXLSGZ17} in a Xilinx Spartan-6 LX75 FPGA chip~\cite{XILINXSP} on the SAKURA-G board~\cite{SAKURA17}. This board is designed for the purpose of evaluating power side channel attacks.
We setup a Tektronix MDO3034 oscilloscope~\cite{TEKMDO}, with a sampling frequency of 2.5GHz, to acquire the power trace from the FPGA board. 



For passive and active adversaries, we propose attack methods for them separately. The whole attack flow is illustrated in Fig.~\ref{fig:overview}. In the first step, we collect the power traces of the FPGA when it performs the convolution with different kernels. Then we adopt an extraction algorithm to filter out noise and get the real power consumption, whose details will be shown in Section~\ref{sec:power_extract}.
After the power extraction stage, passive adversaries try to locate pixels belonging to image background from the extracted power.
Then the silhouette of foreground objects is revealed. The details of this attack are shown in Section~\ref{sec:background_detection}.
For active adversaries, before the actual attack, they build a ``power template''~\cite{RechbergerO04} using the power measured with different kernels and the input image.
The power template exploits the relations between power consumption and pixel values and can generate a set of pixel candidates when queried with power consumption in actual attacks.
The final step for active adversaries is to recover the image by selecting the best pixel candidate from the generated set. Section~\ref{sec:power_templates} introduces the power template attacks in detail.

We conducted experiments with images in the MNIST dataset~\cite{MNIST}, a dataset for handwritten digits recognition. We try to recover the image with both background detection and power template, shown in Fig.~\ref{fig:mnist_all_digits} in Section~\ref{sec:appendix-mnist} of the Appendix.
For images from background detection, the general shape of the original image is retained while the images recovered with power template retain more details and they are more similar to the original images in visual effect.

\section{Power Extraction}\label{sec:power_extract}
Ideally, the power collected from the oscilloscope is periodic and its period shall be same with the clock signal, as the internal activity is triggered by the clock pulse. The power trace in one period shall reflect the total power consumed in this cycle. However, this assumption is not valid due to noises and distortions in the power collecting procedure. Some of the noise sources can be modeled as a capacitor-based filter system which blends power consumption of neighboring cycles and thus makes the raw power trace inaccurate for pixel inference.
In this section, we present an efficient method to extract real power consumption from the noisy and distorted power traces.

\subsection{Interference Sources}

\begin{figure}[tbph]
  \centering
  \includegraphics[width=0.94\linewidth]{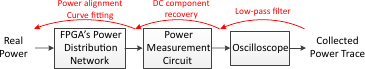}
  \caption{Power measurement path in our experiment setting.}
  \label{fig:power_extract_diagram}
\end{figure}

We illustrate three critical components on the power measurement path in Fig.~\ref{fig:power_extract_diagram}. Driven by clock pulses, CMOS transistors in the FPGA used for computation become active and draw current from power supply. The current is delivered through the \emph{power distribution network} which leads to a voltage drop in the \emph{power measurement circuit}. The voltage drop is then captured by the \emph{oscilloscope's probe} placed on the power supply line and recorded as the power trace.

\begin{figure*}
  \centering
  \includegraphics[width=1.0\linewidth]{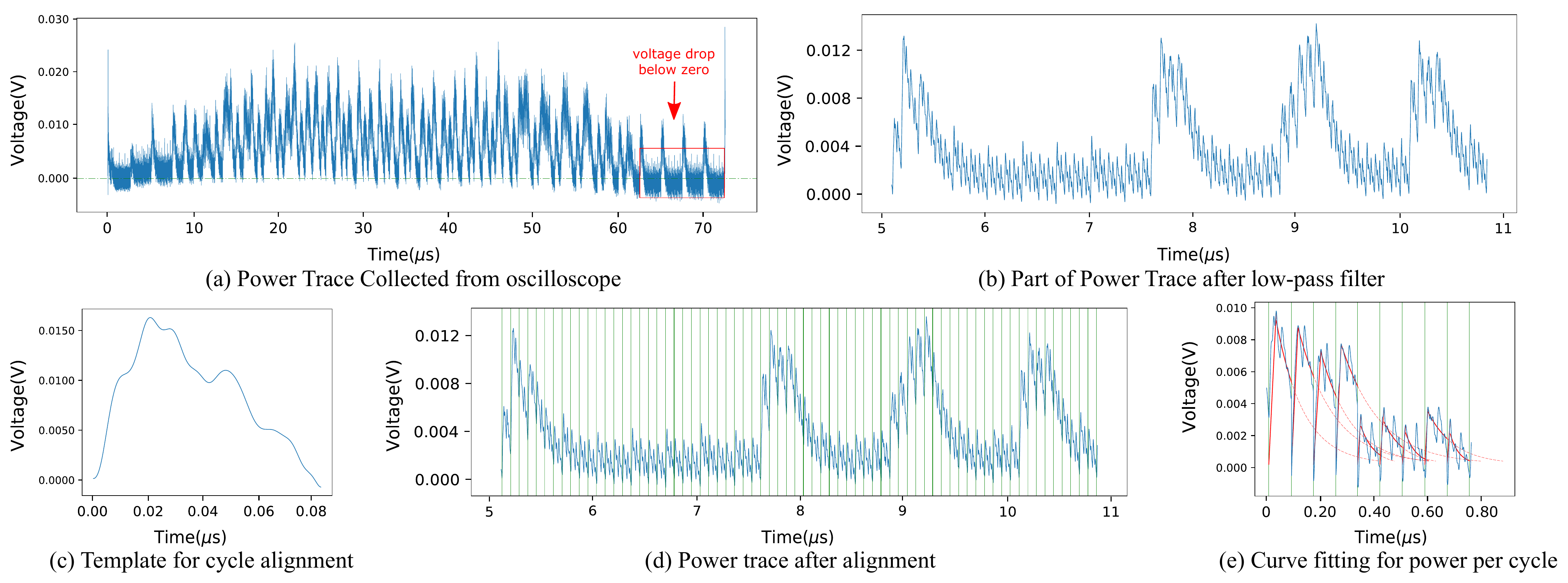}
  \caption{Power extraction on traces collected.}
  \label{fig:power_process}
\end{figure*}

All three components incur certain kinds of interference on the measured power signals. The noise in the measurement of oscilloscope is white noise, which mainly comes from environmental fluctuations. 

The adopted FPGA board~\cite{sakuracir} offers two options for power measurement: we can either directly measure the raw voltage on the resistor or the amplified signal through an amplifier.
It is crucial for the success of our attack to use the amplified signal as the raw voltage on the resistor is about several millivolts, which is around the same level with noise.
However, the amplifier circuit is only able to amplify the AC components of the power traces, which results in the voltage drop below zero at the end of the power traces, as illustrated in Fig.~\ref{fig:power_process} (a). 
The drop not only induces inaccuracies when we recover the power for each cycle, it also hinders correct curve-fitting procedure in subsequent procedures.
We analyze the frequency response of the power measurement circuit with a circuit simulator~\cite{NI}, and find that the whole circuit behaves like a high-pass filter with the cut-off frequency of around 250Hz.

The exact effect of FPGA's power distribution network on power signals is hard to model as we are not knowledgeable about the design details, but we assume it can be regarded as an RC filter. This is because power distribution network is often in a tree-like shape and implemented with metal wires. The distributed wire's resistance and the inter-wire capacitance can be regarded as an RC filter in a lumped model.

\subsection{Extraction Methods}
For the noise in the oscilloscope measurement procedure, we use low pass filters to eliminate them. 
For the distortion from the RC filters, 
though techniques directly reversing the distortion effect exist, they are very sensitive to small deviation~\cite{o1997pragmatic} in the original signal.
Thus, they are not applicable to the power traces from noisy channels.
We propose to solve the problem by analyzing the approximate effect of the RC filters with two dedicated methods: DC component recovery, power alignment and curve fitting.

\noindent
\textbf{Low-Pass Filter: }
For the noises induced in the oscilloscope measurement, 
a low-pass filter is enough by filtering out most high-frequency noises as the low-frequency noises are small compared to useful signals. We apply a filter whose cut-off frequency is 60MHz to the acquired power traces and 
the result is shown in Fig.~\ref{fig:power_process} (b).

\noindent
\textbf{DC Component Restoration: }
For the distortion induced by the power measurement circuit, we propose to recover the DC component. From the simulation result, the cut-off frequency of equivalent high-pass filter (250Hz) is far lower than the accelerator's working spectrum (more than 15kHz, as the total running time is around 70$\mu$s).
So only the DC component is filtered by the power measurement circuit. 
To recover it, we obtain the discrete time impulse response of the power measurement circuit via simulation as follows:
\begin{equation}
h(n) =
\begin{cases}
0 & \quad \text{if } n < 0 \\
1 & \quad \text{if } n = 0 \\
-\frac{T}{\tau}e^{-\frac{nT}{\tau}} & \quad \text{if } n > 0 \\
\end{cases}
\end{equation}
wherein $T$ stands for the sampling interval and it is 0.4ns in our case. $\tau$ represents the time constant, which is the reciprocal of the angular cutoff frequency $\tau = \frac{1}{2\pi f}=640\mu s$. So we propose to recover the original power trace by reversing effects of the power measurement circuit, which can be modelled as $x(n) = r(n) * h(n)$, using the following formula:
\begin{equation}
r(n) = x(n) - \sum_{i=0}^{n-1}x(i)h(n-i)
\end{equation}
wherein the $x(n)$ represents power samples collected while the $r(n)$ stands for sample points in the recovered trace.

\noindent
\textbf{Power Alignment and Curve Fitting: }
Though the FPGA's power distribution network is also RC-filter-like, it is hard to approximate it to simple low-pass or high-pass filters as its frequency response overlaps the spectrum of power traces.
Alternatively, based on this RC filter assumption, we further assume the power trace acquired per cycle is similar to the capacitor's charging and discharging curve. Then we use curve fitting tools to obtain the exact power consumption in one cycle.
In the first step, we need to align the power trace with the clock signal. A template signal, representing a typical power trace in one clock cycle as shown in Fig.~\ref{fig:power_process} (c), is carefully chosen from the filtered power trace manually and we calculate the \emph{Pearson correlation coefficient} of the template signal with each sample point on the original power trace. We choose the points with maximum coefficients to be the alignment points. The aligned power trace is shown in Fig.~\ref{fig:power_process} (d).

For power signals in each cycle, they all rise sharply at first and then gradually descend, which comes from the charging and discharging of the equivalent capacitor in power distribution network.
Thus, we fit the power curve with capacitor's charging formula $V_c(t)$ and discharging formula $V_d(t)$ as follows:
\begin{equation}
  V_c(t) = V_p(1-e^{-\frac{t}{RC}}), V_d(t) = V_pe^{-\frac{t}{RC}}
\end{equation}
in which the $V_p$ represents the final voltage at the charging stage and the initial voltage in the discharging phase. $RC$ is the product of equivalent resistance and capacitance of the power distribution network, also known as RC time constant, represented by $\tau$. The whole power extraction algorithm is listed in Algorithm~\ref{alg:power_extraction}. Also we illustrate this procedure in Fig.~\ref{fig:power_process} (e). The algorithm is run cycle-wise: for each cycle, we estimate optimal $V_p$ and $\tau$ from the power trace using curve fitting function and calculate the trailing power in subsequent cycles.
The final power for current cycle is accumulated by the power in this cycle and the trailing power.
The trailing power is then subtracted from following power traces. The computation continues until all aligned cycles are processed.
The solid red line in Fig.~\ref{fig:power_process} (e) shows the optimal curve we find while the dash red line shows the trailing power for each cycle.

\begin{algorithm}[tbhp]
\KwIn{A vector $p$ containing all sample points in collected power trace;
A set of tuples indicating start and end indexes for each cycle in the power trace $\mathbf{S} = \{(i_{st}, i_{ed})_j | \text{for } j \text{ in all cycles}\}$
}
\KwOut{A vector $P$ containing the real power consumption for each cycle}
\For{$j$ in all aligned cycles} {
    $i_{st}, i_{ed} \leftarrow \mathbf{S}[j]$\;
    $p_{trace} \leftarrow p[i_{st}:i_{ed}]$\;
    $param \leftarrow \text{CurveFitting}(p_{trace})$\;
    $p_{trail} \leftarrow \text{GenerateTrailingTrace}(p_{trace}, param)$\;
    $P[j] \leftarrow sum(p_{trace}) + sum(p_{trail})$\;
    $p[i_{ed}:i_{ed}+len(trailing)] -= p_{tailing}$\;
}
\caption{{\sc Cycle Power Extraction}}
\label{alg:power_extraction}
\end{algorithm}

\section{Background Detection}\label{sec:background_detection}
In this section, we first discuss the intuition of our background detection attack. Then we introduce the threshold-based attack method and at last we evaluate it with MNIST datasets~\cite{MNIST}.
\subsection{Intuition}
For passive adversaries, the intuition to attack the DNN accelerator comes from the power model: the power consumption is determined by the internal activities, especially by those in the convolution unit which takes the largest portion of power consumption. If the data inside the convolution unit remain unchanged between cycles, the internal transitions induced are limited. Thus, the power consumption shall be small.
Based on this insight, by observing the magnitude of power consumption in each cycle, passive adversaries can determine whether the related pixels share similar values. These similar pixels most probably belong to the pure background of the image.
As a result, the silhouette of the foreground object naturally revealed by locating all pixels belonging to background and the privacy of user's information may be infringed via adversaries' visual inspection.

Though many real-world images have a messy background, some types of privacy-sensitive images happen to contain pure background, such as medical images from ultrasonography or radiography. If the adversaries are able to recover the shape of foreground object, they may be able to identify the organ being scanned and thus infer the health condition of a particular patient.  

\subsection{Attack Method}
The basic idea of the attack is to find a \emph{threshold} to distinguish cycles processing background pixels based on the magnitude of power consumption.
However,
deciding the exact threshold is not a trivial task as we cannot observe a clear gap in the distribution of power consumed in each cycle, as shown in histogram in Fig.~\ref{fig:bck_power} (a). 
We assume the power consumed in cycles processing foreground pixels are evenly distributed across a large range while the power consumption of rest cycles aggregates at the bins of smaller values.
So we are expected to observe a peak in cycle counts for smaller cycle power consumption. In this case, we decide the threshold by finding the maximal decrease in cycle count:
\begin{equation} 
  P_T = \argmax_{P} C(P - B) - C(P) 
\end{equation}
wherein the $C(\cdot)$ is the function returning the cycle count for a particular power consumption, $B$ is the bin size.

After the threshold is determined, we filter out all cycles whose power consumption is above the threshold. Then we locate all corresponding pixels for the left cycles. These pixels are regarded as background pixels and then we can get a black-and-white image for further examination and analysis.

\subsection{Evaluation}\label{subsec:background_evaluation}
\noindent
\textbf{Experiment Setup: }
We performed our attack on the CNN accelerator used to classify the digits in MNIST datasets.
The size of the images in MNIST datasets is 28$\times$28. The images have a clear black background which satisfies the pre-requisite of our background detection method. 
For the CNN accelerator~\cite{ZhaoSZXLSGZ17}, we set the line size of the line buffer to 28, input channel to 1 and the kernel size to 3$\times$3 and 5$\times$5. We adopted two models for experiment with their details shown in Table~\ref{tbl:mnist_models}. The only difference between the two models is the kernel size as it directly determines the number of pixels involved in the convolution unit and affects the granularity of recovered images.

\begin{table}[htbp]
\centering
\caption{Binarized CNN Model Details on MNIST datasets}
\label{tbl:mnist_models}
\begin{tabular}{|l|c|c|}
\hline
  & Model 1 & Model 2 \\
\hline
No. of layers & \multicolumn{2}{c|}{4}   \\
\hline
Accuracy on testing sets & 99.42\% & 99.27\% \\
\hline
Type of 1st layer & \multicolumn{2}{c|}{Convolution} \\
\hline
Kernel size in 1st layer & \textbf{3 $\times$ 3} & \textbf{5 $\times$ 5} \\
\hline
No. of kernels in 1st layer & \multicolumn{2}{c|}{64} \\
\hline
\end{tabular}
\end{table}

We synthesized the CNN accelerator design to FPGA and loaded the model parameters into the accelerator before the inference stage. 
We randomly chose 500 images from the MNIST testing set to evaluate our attack method. 
Both models contain 64 different kernels in the first layer and for each kernel, we recorded the power trace when the accelerator performed the convolution in the first layer. As our algorithm recovers the pixel values on a cycle base, we need to precisely identify the power trace fragment for the convolution in the first layer. It is trivial to locate the start point on the power trace and the length of fragment can be determined from the total clock cycles needed to finish the convolution computation.

\noindent
\textbf{Evaluation Metric: }
We evaluate the quality of recovered images with two metrics: \emph{pixel-level accuracy} and \emph{recognition accuracy}.
Pixel-level accuracy is to evaluate the precision of our attack algorithm and it is defined as follows: 
\begin{equation}
\alpha_{pixel} = \sum_{\mathbf{x} \in \mathbf{I}}\llbracket f(\mathbf{x}) = marker(\mathbf{x}) \rrbracket / |\mathbf{I}|.
\end{equation}
in which $\mathbf{x}$ represents a pixel in the targeted image $\mathbf{I}$. $f(\mathbf{x})$ means the background marker (whether it belongs to background) predicted by our algorithm while $marker(\mathbf{x})$ represents the correct marker.
For MNIST images, we regard all pixels with value 0 (i.e., pure black pixel) as the background pixels.
We also evaluate the cognitive quality of the recovered image with recognition accuracy. We feed every recovered image to a high-accuracy MNIST classification model and compare the prediction result with its correct label. In the following experiment, we use a multi-layer perceptron network~\cite{HubaraCSEB16} with an accuracy of 99.2\% as a golden reference to evaluate the cognitive quality.

\begin{figure}
  \centering
  \includegraphics[width=1.0\linewidth]{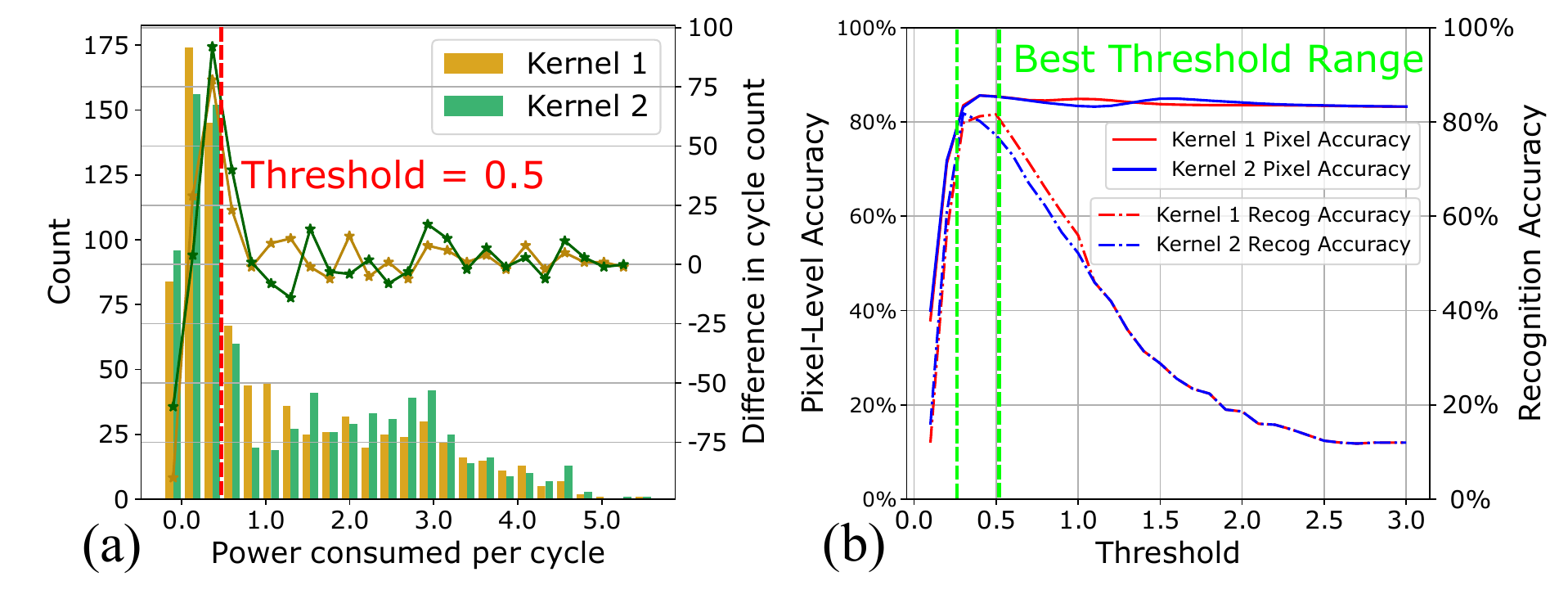}
  \caption{(a) Histogram of power consumed per cycle with two kernels. (b) Accuracy with two kernels on model 1.}
  \label{fig:bck_power}
\end{figure}

\noindent
\textbf{Choices of Threshold and Kernel:}
We show a histogram of power consumed in each cycle in Fig.~\ref{fig:bck_power} (a). 
In the figure, we draw the histograms for the power computed with two different kernels from model 1 and they manifest similar trends: with the increase of power consumed per cycle, 
the cycle count rises at first and then descends sharply at the value of 0.5. After that, the cycle count gradually decreases and finally reaches 0. Based on the threshold selection criteria, we choose the threshold at 0.5 for this image.

To demonstrate the importance of threshold choice in the attack, we recover the silhouette images using various threshold values from 0.1 to 3.0 with a step size of 0.1 and illustrate the two metrics in Fig.~\ref{fig:bck_power} (b).
The pixel-level accuracy is drawn with solid lines while the recognition accuracy is drawn using the dot-dash lines. 
We observe that as the threshold increases, the pixel-level accuracy increases to its peak value around 85.6\% for both kernels at first and then it gradually decreases to 83.3\%.
The recognition accuracy for these two kernels also follows similar trends: they first rise to its peak accuracy,
but they drop significantly as the threshold increases. 
For kernel 1, it reaches its peak value 81.6\% at threshold 0.5 while for kernel 2, it reaches peak of 81.8\% at threshold 0.3.

When the threshold approaches its optimal value, more and more background pixels are correctly identified, so we observe an increase in both pixel-level accuracy and recognition accuracy. After the threshold exceeds the optimal value, the pixel-level accuracy only drops a little, while the recognition accuracy falls remarkably. 
This is because the actual number of foreground pixels is smaller than the background pixels. When threshold increases, the number of misclassified pixels is limited thus it does not affect the pixel-level accuracy much. However, as foreground pixels are key components to construct digit strokes, 
their misclassification leads to significant loss in recognition accuracy.
According to these two curves, the optimal range of threshold lies  between [0.3, 0.5], which is the region in two green dashed lines in Fig.~\ref{fig:bck_power} (b). It is consistent with the threshold selection criteria we raised in last subsection.
Another conclusion from the experiment is that both accuracy metrics are seldom affected by the choices of kernel. 
Therefore, adversaries can acquire acceptable images by only attacking on the power trace for one kernel.


\begin{figure}
  \centering
  \includegraphics[width=0.85\linewidth]{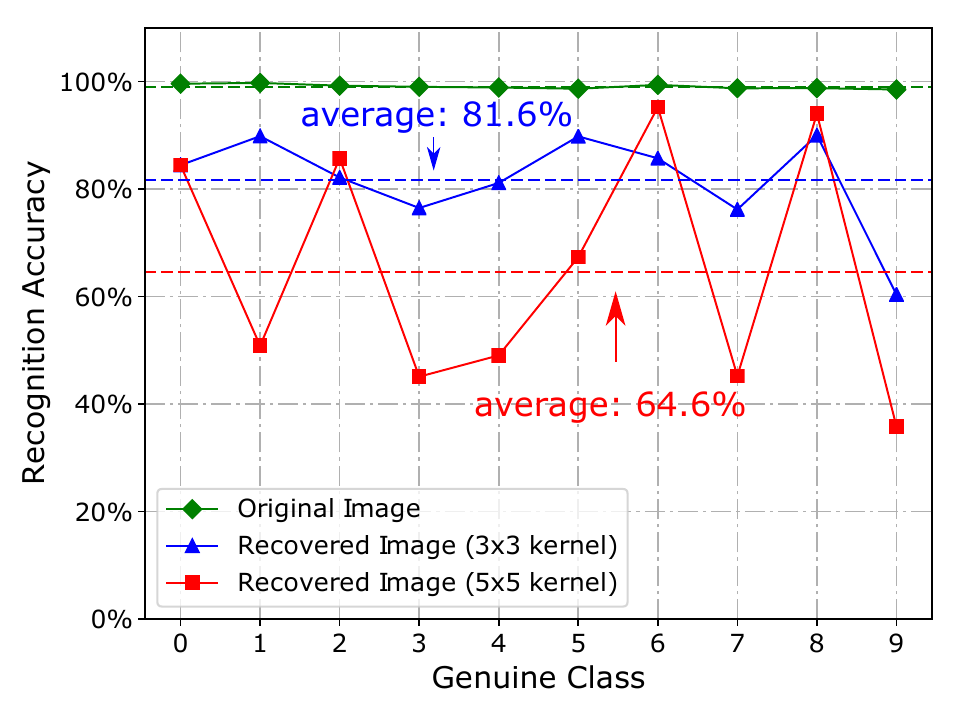}
  \caption{Recognition accuracy for model 1 and model 2.}
  \label{fig:bck_class_per_digit}
\end{figure}

\noindent
\textbf{Kernel Size:}
Kernel size can be a significant factor affecting both the pixel-level and recognition accuracy.
The average pixel-level accuracy for model 1 (3$\times$3 kernel size) is 86.2\% while the accuracy for model 2 (5$\times$5 kernel size) is around 74.6\%. The recognition accuracy is shown in Fig.~\ref{fig:bck_class_per_digit}: on average case, 81.6\% for images recovered from power acquired with model 1 and 64.6\% with model 2. 
The accuracy degradation comes from the information loss when kernel size increases.
It is because our algorithm is only able to find cycles that deal with background pixels via thresholding, which requires all the pixels inside the convolutions units to be identical. In other words, if the convolution unit contains a non-background pixel, all other background pixels may be mis-identified as foreground pixels by our proposed algorithm.
This effect is similar to the morphological dilation operation~\cite{rafael2002digital} in the digital image processing which widens the shape of foreground objects. The recovered image looks ``fatter'' than the original image in visual effect.
Key structures smaller than the kernel size are more probable to disappear, resulting in the degradation of recognition accuracy. 

Another discovery is that the recognition accuracy is various for different digits, especially for model 2. Fig.~\ref{fig:bck_class_per_digit} shows the recognition accuracy categorized by the class of digits. The accuracy of classifying original image is almost the same and nearly perfect for all classes. The recognition rates of digits 3, 7 and 9 are below average for images recovered with model 1. Meanwhile, the accuracy of digits 1, 3, 4, 7 and 9 drop significantly for model 2.
We consider the discrepancy among different digits comes from inherent structure of digits and equivalent dilation effect of recovered image.
For example, the image of digits 1 recovered from model 2 is much ``fatter'' than that from model 1 due to dilation effect, so it is more probable for the classification network to misclassify it as digits 8, causing a low recognition rate.

\begin{figure}
  \centering
  \includegraphics[width=0.95\linewidth]{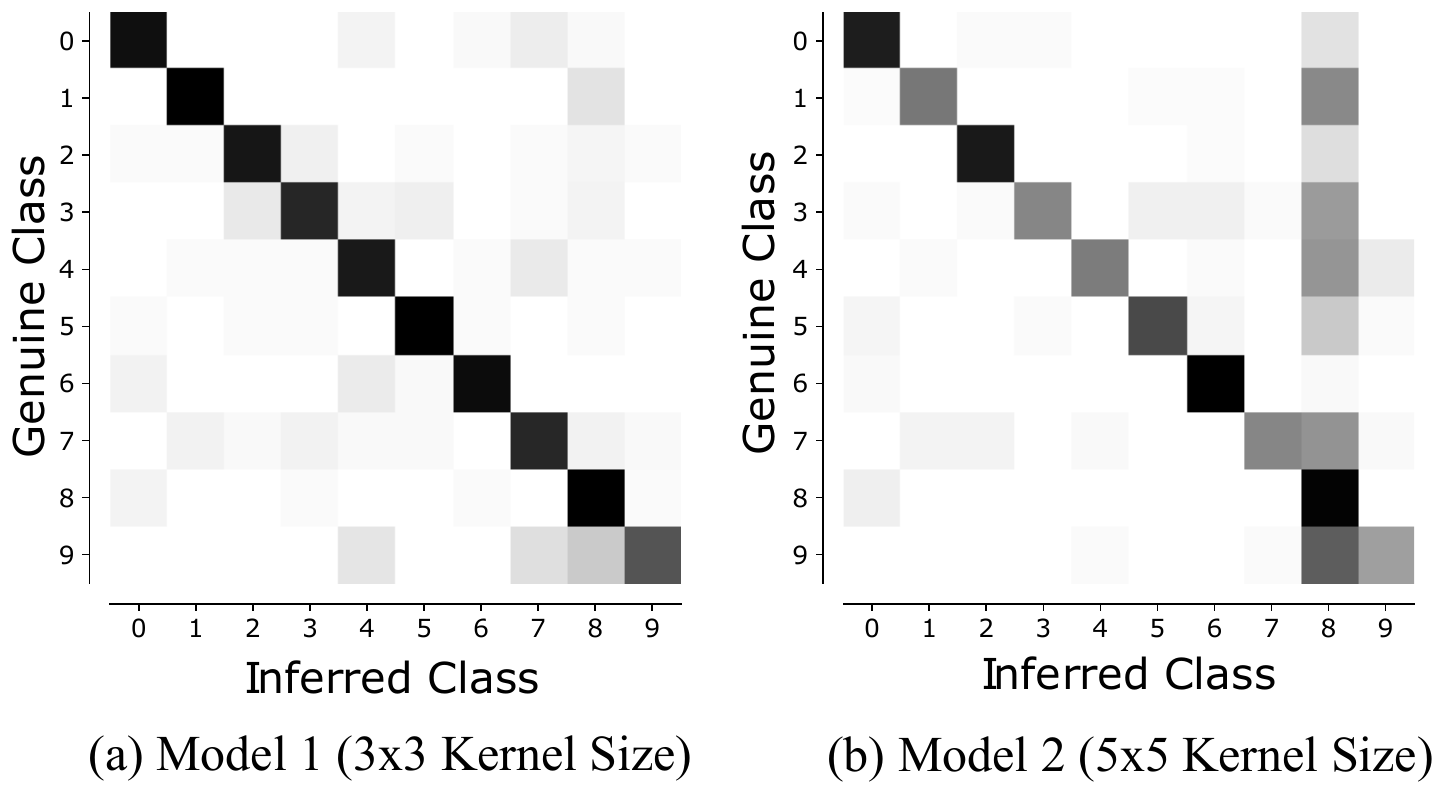}
  \caption{Classification map for model 1 and model 2.}
  \label{fig:misclass_map}
\end{figure}

We also investigate the classification result of recovered images for each digit with both models.
The results are drawn in the classification map shown in Fig.\ref{fig:misclass_map}. Each cell $(i, j)$ in the map represents the portion of images with correct class $j$ which are predicted as class $i$, wherein the portion is illustrated with the darkness.
For both models, the darkest color all lies on the diagonal of the map, which means the classification network is able to correctly predict in most cases. 
We observe the recognition accuracy of digits 8 is quite high (around 90\%) for model 2 in Fig.~\ref{fig:bck_class_per_digit}. However, the precision is not. From Fig.~\ref{fig:misclass_map} (b), the cells in the column of inferred digits 8 are darker than other cells in the same row except for genuine class 8. 
Thus, for an image inferred as digit 8 may have larger probability to be other digits actually because of its low precision. This is because the inherent shape of 8 is large than other digits and the classification network is more inclined to classify a ``fatter'' image, which is caused by dilation effect, to digits 8.


To conclude, the kernel size affects both the pixel-level accuracy and recognition accuracy due to their equivalent dilation effect induced by kernels. The recognition accuracy of different digits also varies because of their inherent structure.

\noindent
\textbf{Complexity: } The attack method only attacks one power trace. As the power extraction and background detection procedure are cycle-based, the time complexity is proportional to the total number of cycles to compute the convolution which is determined by the image size $O(S_x \times S_y)$, where the $S_x$ and $S_y$ is the length of the image in two dimensions.
The total time used is short in practice. It takes around 6s to obtain one power trace and 5.7s for power extraction. For actual image reconstruction it only takes 0.01s.


\section{Image Reconstruction via Power Template}\label{sec:power_templates}
In this section, we propose an attack method, for active adversaries, to recover the details of images used in the inference process. Instead of predicting background marker, we try to obtain values for each pixel.
The section is organized similarly with Section~\ref{sec:background_detection} with three subsections: intuition, attack method and evaluation.

\subsection{Intuition}
The search space to recover pixel values is prohibitively large even if only considering the pixels in a small local region.
Suppose the targeted model uses a 3$\times$3 kernel size for the first convolution layer, the number of pixels involved in the convolution in one cycle is 12 (see the analysis in Section~\ref{subsec:pixel_attack}). Typically a pixel can have a value ranging from 0 to 255, so the total combinations for the pixels involved is around $256^{12} \approx 7.9\times10^{29}$. 
Iterating all combinations of pixel values in brute force is inefficient to perform the attack.
Thus, for active adversaries, we propose to reduce the search space significantly by building a ``power template''. 
As active adversaries are able to profile the relationship between power consumption with arbitrary input images, the pre-built ``power template'' is able to efficiently predict the pixel value at actual attack with knowledge acquired at profiling stage. 


As illustrated in the third part of Section~\ref{sec:background},
the power consumed in each cycle is determined by the data inside convolution unit, which comprise pixel values and kernel parameters.
Typically, the same inputs are convolved with different kernels in the convolutional layer. For a specific region of pixels processed by convolution unit, we can regard the power consumption acquired from different kernels as a unique feature to infer the value of the pixels.
Based on this intuition, we build a ``power template'' storing the mapping of power consumption to pixel values so that adversaries can 
produce a set of possible pixel values from a vector of power consumption retrieved at attack time. 
Finally, after we acquired many candidate values for each pixel from multiple power vectors, an image reconstruction algorithm is adopted to select the best candidate. As one pixel is processed in multiple cycles, the target of the selection is to find candidates in these cycles predicting similar values at this pixel.



\subsection{Attack Method}\label{subsec:pixel_attack}
In this subsection, we introduce the detailed steps to recover pixel values of the input image.
First we discuss how to build the power template at the profiling stage. Then, with the extracted power from different kernels acquired at attack time, we demonstrate the method to get candidate pixel values from power template. Finally we present an algorithm to reconstruct the image from these candidates.

\noindent \textbf{Power Template Building: }
Power template stores the mapping between pixel values and its corresponding power consumption when convolved with different kernels.
In the profiling phase, for each input image, we collect multiple power traces from the FPGA loaded with different kernels and obtain power consumption at each cycle using power extraction in Section~\ref{sec:power_extract}.

The power consumed in each cycle is determined by the state transitions of the convolution unit. The kernel remains constant between cycles, while the pixels are shifted within a row. So the number of related pixels in one cycle is $K \times (K + 1)$ when the kernel size of is $K \times K$. For example, suppose at cycle $j-1$, pixels from position $(x, y)$ to $(x + K, y + K)$ are inside the convolution unit, while at cycle $j$, pixels from position $(x, y + 1)$ to $(x + K, y + K + 1)$ are processed. The power consumed in cycle $j$ is induced by change in the convolution unit, so all pixels from $(x, y)$ to $(x + K, y + K + 1)$ determine the power consumption. We represent the this region as $\Omega_j$ and the pixel values in this region as $\mathbf{Px}_j=\{pv(\mathbf{x}) \text{ } | \text{ for }\mathbf{x} \in \Omega_j\} $. These pixel values are named \emph{related pixels for $j$-th cycle}. Further, we represent the power consumption in $j$-th cycle when the image is computed with $i$-th kernel as $\rho_{i, j}$.

So for each cycle, we obtain the related pixels $\mathbf{Px}_j$ and the power collected with different kernels, namely \emph{power feature vector}, represented as $\boldsymbol{\rho}_j = (\rho_{0, j}, ..., \rho_{|K|, j})$. $|K|$ is the number of kernels used.
For one input image at profiling stage, the power template is constructed by adding all pairs of related pixels and corresponding power feature vector for all cycles. We define the power template formally as follows:
\begin{equation*}
PT = \{(\mathbf{Px}_j : \boldsymbol{\rho}_j) | \text{for }j\text{ in all cycles}\}.
\end{equation*}
The final power template is the union set of power templates constructed by every input image.


\noindent \textbf{Candidates Generation: }
Based on the assumption that similar pixels processed in the convolution unit generate similar power feature vector, the straight-forward way to get pixel candidates is to find pixel values in the power template whose corresponding power feature vector is closest from that extracted during attack. However, this method easily fails due to limited samples enrolled in the power template.
Hence, we propose to divide the power feature vector into several groups and search them in the power template respectively.
After we get the pixel candidates for each group, we take the intersection of them to generate the final candidate set for image reconstruction.

To be specific, for a specific cycle $j$, we acquire a power feature vector $\boldsymbol{\rho}'_j$ from measured power traces and separate them into several groups of same size $\boldsymbol{\rho}'_j = \{{\boldsymbol{\rho}_j^1}', ..., {\boldsymbol{\rho}_j^m}'\}$.
For each entry in the power template, we do the same separation, i.e., $\boldsymbol{\rho} = \{\boldsymbol{\rho}^1, ..., \boldsymbol{\rho}^m\}$.
For each group of vectors ${\boldsymbol{\rho}_j^m}'$, we search the same group of power feature vectors in the power template $\boldsymbol{\rho}^m$ and return the related pixels if the distance between two groups is within a threshold $\delta$.
The candidate set, consisting of all returned related pixels, is given by
\begin{multline*}
S_m^j = \{\mathbf{Px} \text{ }| \\
\text{for all } (\mathbf{Px} : \boldsymbol{\rho}) \text{ in } PT \text{ and } dist(\boldsymbol{\rho}^m, {\boldsymbol{\rho}_j^m}') < \delta\}
\end{multline*}
where the distance metric is defined by
\begin{equation*}
dist(\boldsymbol{\rho}^m, {\boldsymbol{\rho}_j^m}') = \sum_{i \in K_m}||\rho_{i} - \rho_{i, j}'||_2,
\end{equation*}
The $K_m$ represents the kernel indexes of power features grouped to the $m$-th group.
The final candidate set for the specific cycle $j$ is given by the intersection of the candidate sets from all different groups, i.e., $S^j = \cap_m S_m^j$.

\noindent \textbf{Image Reconstruction Algorithm:}
After obtaining the candidate sets for all cycles, we have many possible values for each pixel and the target is to find the closest one to the actual value.

As we have noted that the same pixel is processed in different cycles, so the candidates selected among these cycles shall be consistent on the value of the same pixel. We use this as a criterion to find the optimal selection.
Suppose for a pixel at position $\mathbf{x} = (i, j)$, there are t cycles processing this pixel: $C_{\mathbf{x}} = \{c_1, ..., c_t\}$. The candidate sets for these cycles are $S^{c_1}, ..., S^{c_t}$. From each candidate set, we choose a candidate with a selector $Sel_{c_t}$ as $\mathbf{Px_{c_t}} = S^{c_t}[Sel_{c_t}]$, and we find the pixel value $pv(\mathbf{x})$ at position $\mathbf{x}$ in $\mathbf{Px_{c_t}}$. The variance of pixel values at position $\mathbf{x}$ selected from different candidate set shall be small.
The objective of our image reconstruction algorithm is to find a selector vector so that the selected candidates minimize the sum of the variance of all pixels in the image. It can be represented as follows:
\begin{equation}
\underset{Sel}{min} \sum_{\mathbf{x} \in \mathbf{I}} var(\{pv(\mathbf{x}) \text{ } | \text{ } pv(\mathbf{x}) \in \mathbf{Px_{c_t}} \text{ for } c_t \in C_{\mathbf{x}}\})
\label{eq:opt_targ}
\end{equation}
After the selector vector is determined, for each cycle, we get only one candidate. But for each pixel, we get multiple candidates from cycles processing it. To get the final value of the pixel, we take the average of these candidates.

This optimization problem is not easy to solve, here we present a greedy heuristic method 
shown in Algorithm~\ref{alg:image_recover}. 

\begin{algorithm}
\KwIn{Sets of candidates for related pixels $\{S^1, ..., S^T\}$ for all $T$ cycles extracted
}
\KwOut{A selector $Sel$ of size $T$ that minimizes the sum of variance defined in Eq.~\ref{eq:opt_targ}}
TempStore = []\;
$S_r \leftarrow \text{Random}(\{S^1, ..., S^T\})$\;
\ForEach {related pixels $\mathbf{Px}$ in $S_r$} {
    $\mathbf{I} \leftarrow \text{EmptyImage()}$\;
    $Sel \leftarrow \text{EmptyVector()}$\;
    \ForEach {pixel $pv(\mathbf{x})$ in $\mathbf{Px}$} {
        $\mathbf{I}[\mathbf{x}] \leftarrow pv(\mathbf{x})$\;
    }
    \While {there exists candidate set unprocessed} {
        $t \leftarrow \text{SelectCycle}(\mathbf{I})$\;
        $\mathbf{Px'} \leftarrow \text{MinimalDiscrepancy}(S^t, \mathbf{I})$\;
        $\mathbf{I} \leftarrow \text{Update}(\mathbf{I}, \mathbf{Px'})$\;
        $Sel[t] \leftarrow \text{Idx}(\mathbf{P_x'})$\;
    }
    $\sigma \leftarrow \text{CalculateVariance}(Sel, \{S^1, ..., S^T\})$\;
    TempStore.add(($\sigma$, $Sel$))\;
}
$Sel \leftarrow \text{FindMinimal}(\sigma, \text{TempStore})$\;
return $Sel$
\caption{{\sc Image Reconstruction Algorithm}}
\label{alg:image_recover}
\end{algorithm}

In Algorithm~\ref{alg:image_recover}, we start with a random candidate set and for each candidate in the set, we initialize an empty image with the pixels in the candidate (Line 6--8), other pixels are left undecided.
Then we greedily search the unprocessed candidate set and find the candidate whose $\Omega_t$ overlaps current image to the largest extent (Line 10) and who has the smallest distance with the overlapped pixels in the current image (Line 11).
The image is then updated accordingly with the candidate and its index is recorded (Line 12--13) .
This process is repeated until all candidate set is processed, and then a selector of size $T$ is generated.
We calculate the variance defined in Eq.~\ref{eq:opt_targ} for the selector (Line 15--16). After all candidate in the original set $S_r$ is processed, we find the selector with minimal variance and return it as the final result (Line 18--19).

\subsection{Evaluation}

\noindent \textbf{Experiment Setup: }
We followed the same experiment setup in Section~\ref{subsec:background_evaluation} except that we used 300 images to build the power template and the left 200 images to evaluate attack method. The pixel value in the MNIST digits is in the range of [0, 255].
We chose to build the power template with power traces collected with 9 different kernels instead of all 64 kernels, because it already provides enough precision to recover the input image.

\noindent \textbf{Evaluation Metric: }
We use the same evaluation metrics with those in the background detection except the pixel-level accuracy is re-defined with pixel values instead of background markers as follows:
\begin{equation}\label{eq:pixel_dist}
\alpha_{pixel} = \sum_{\mathbf{x} \in \mathbf{I}} ||pv(\mathbf{x}) - pv_g(\mathbf{x})||_2 / |\mathbf{I}|
\end{equation}
in which $pv$ represents the pixel value in the recovered image while $pv_g$ means the pixel value in the original image.

\noindent \textbf{Candidates Generated:}
To evaluate the effectiveness of grouping power vectors in the power template,
we list the statistics of candidates returned by the power template in Table~\ref{tbl:candidates}. As we collected power traces from 9 different kernels, so the length of power feature vector for each cycle is 9. In the experiment, we divided the power features into 4 groups of size 2 (using first 8 features) and 3 groups of size 3 respectively. 
The number of candidates returned by the power template for one group is denoted as $|S|$ and the distance threshold is $\delta$. 
We also calculate the distance between each candidate and the genuine related pixels and represent the minimal distance as $D_{min}$, which serves as the quality metric of returned candidate set: the smaller, the better.
For the final candidate set, i.e. the intersection of all candidate set from different groups, we also report the number of candidates in the set $|S^{\cap}|$ and the minimal distance $D_{min}^{\cap}$. Table~\ref{tbl:candidates} shows the average of these numbers among all cycles.

From the table, the number of candidates increases with the increase of threshold $\delta$ as larger search space is included. Also the average of minimal distance decreases when more candidates are included. For smaller $\delta$, such as 0.1 and 0.2, the number of candidates returned are small, and for many cycles, we are not able to find a match inside the template. Thus, smaller $\delta$ may lead to lower precision in finding the related pixels.
For two experiments with different group size investigated, both of them achieves significant reduction in the size of final candidate set, while maintaining similar capability to recover more precise pixels (reflected by small changes of $D_{min}$) at medium or large $\delta$s. 

In all, the grouping of power feature vectors and intersection of candidate sets from power template is effective in reducing the size of pixel candidates for each cycle while maintaining the accuracy at the same time.

\begin{table}[]
  \centering
  \caption{Candidates Statistics for Model 1 (3 x 3 Kernel Size)}
  \label{tbl:candidates}
  \begin{tabular}{|c|c|c|c|c|c|c|c|c|}
  \hline
  \multirow{3}{*}{$\delta$} & \multicolumn{4}{c|}{GroupSize = 2} & \multicolumn{4}{c|}{GroupSize = 3} \\ \cline{2-9} & \multicolumn{4}{c|}{} & \multicolumn{4}{c|}{} \\[-1em]
   & $\overline{|S|}$ & $\overline{D_{min}}$ & $\overline{|S^{\cap}|}$ & $\overline{D_{min}^{\cap}}$ & $\overline{|S|}$ & $\overline{D_{min}}$ & $\overline{|S^{\cap}|}$ & $\overline{D_{min}^{\cap}}$  \\ \hline
  0.1 & 767 & 57 & 107 & 190 & 325 & 153 & 48 & 155 \\ \hline
  0.2 & 1448 & 45 & 351 & 116 & 787 & 90 & 170 & 102 \\ \hline
  0.5 & 3847 & 33 & 1086 & 90 & 2447 & 48 & 715 & 68 \\ \hline
  1.0 & 9457 & 26 & 2223 & 67 & 5890 & 34 & 1571 &  56  \\ \hline
  \end{tabular}
  \end{table}

\noindent \textbf{Image Quality: }
Based on the experimental result in Table~\ref{tbl:candidates}, we proceed the image reconstruction with group size 3 as the final candidate size is relatively small to group size 2. We also determine the $\delta$ to be 1.0 to maintain a high accuracy candidate set for further reconstruction. 
For the left 200 images used for evaluation, using Algorithm~\ref{alg:image_recover}, we recover them from the candidate sets from the power template. We also generate images without using this algorithm for comparison.
Without Algorithm~\ref{alg:image_recover}, for a particular pixel in image,
its value is given by the average of all possible values for this pixel in the returned candidates.
The average pixel-level distance, defined in Eq.~\ref{eq:pixel_dist}, is 1.65 for image generated with Algorithm~\ref{alg:image_recover} and 2.98 for images without it. On an average case, both of them are quite close to the genuine image considering the pixel value range is 0 to 255. This is because the candidates generated from power template are already close to the genuine pixels.

\begin{figure}
  \centering
  \includegraphics[width=0.85\linewidth]{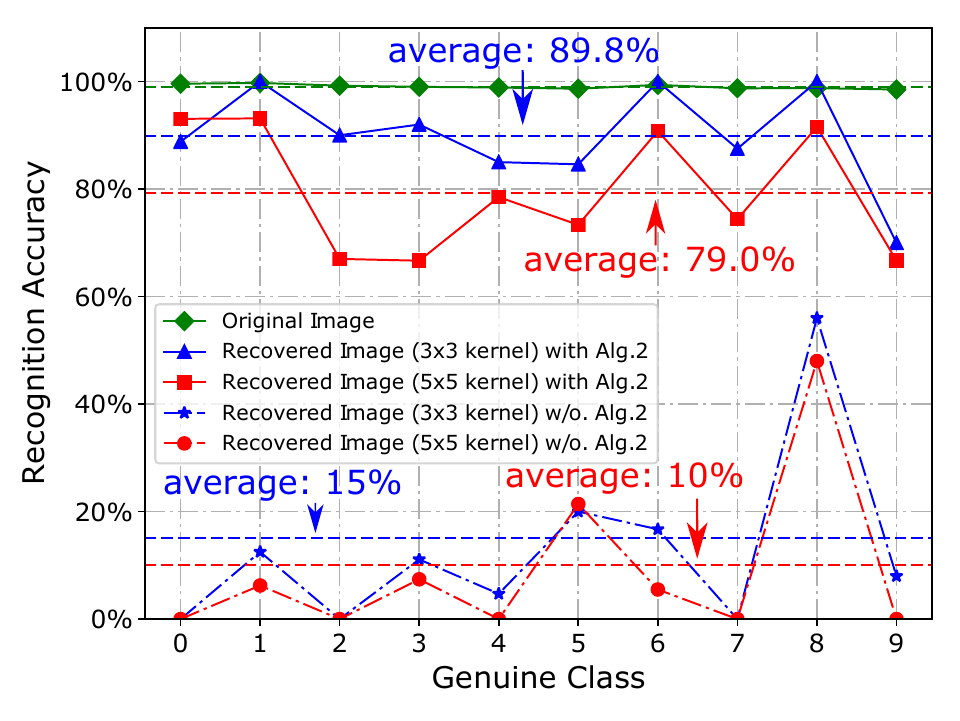}
  \caption{Recognition accuracy for model 1 and model 2 with and w/o. reconstruction algorithm.}
  \label{fig:pwrtem_class_per_digit}
\end{figure}

However, as illustrated in Fig.~\ref{fig:pwrtem_class_per_digit}, the recognition accuracy is much higher with Algorithm~\ref{alg:image_recover}.
The recognition accuracy of images recovered for model 1 (3$\times$3 kernel size) with the algorithm is 89.8\% while the accuracy drops down to 15\% if we take the average of all the pixel candidates.
The same accuracy drop also happens on the images recovered from power trace collected with model 2 (5$\times$5 kernel size), from 79\% to 10\%.
Though the images recovered without the proposed algorithm achieve relatively good pixel-level accuracy, the low recognition accuracy results from its incapability to reconstruct the structure of digits at some critical points, especially at the edge of digits.
On the contrary, Algorithm~\ref{alg:image_recover} considers the consistence of related pixels recovered among cycles, thus it is able to filter out most unrelated pixels.

Finally, enlarging kernel size incurs a little degradation in the recognition accuracy, from 89.8\% to 79\% as more pixels are involved in one cycle so that it is relatively harder to distinguish the genuine pixels.

\noindent \textbf{Complexity: }
We analyze the complexity in three phases: 
The time complexity to build power template is $O(N \times C \times K)$ where $N$ stands for total number of images enrolled, $C$ means the cycles needed to generate one feature map and $K$ is the number of kernels.
The memory complexity in power template building is $O(N \times C \times (K + S))$, where $N \times C$ represents the total amount of entries in power template and $K + S$ is the entry size. $S$ stands for the size of related pixels. 
In the candidate generation, the time complexity of proportional to the size of the power template and the size of returned candidate sets. Finally, for 
the image reconstruction algorithm, 
the most time-consuming part comes from the loop in it (Line 3 -- 17), so its complexity is $O(C \times A^2)$, where $C$ is total number of candidate sets (equal to the cycles needed to generate a output feature map) and $A$ stands for the average size of the candidate set. 
All methods in three phases can be implemented efficiently and we report their running time as follows:
it takes 215.6s to build the power template from 300 images and 157.2s to generate candidates for all cycles in recovering one image. The image reconstruction algorithm costs around 43.2s to finish.
The size of the power template built with 300 images enrolled is around 44MB.

\section{Related Work}\label{sec:related_work}
\noindent \textbf{Neural Network Privacy: }
In~\cite{FredriksonLJLPR14}, authors made a successful attempt to correlate the dosage of certain medicine with a specific patient's genotype from a model used in pharmacogenetics.
Also, on a face recognition system, they managed to reconstruct users' face images enrolled in the training stage from the neural network models~\cite{FredriksonJR15}.
Shokri et al~\cite{ShokriSSS17} presented a membership inference attack to decide whether a particular data record belongs to the model the training set with a black-box access to the model.
Tramer et al~\cite{TramerZJRR16} demonstrated a model inversion attack by exploiting the relationship of queries and confidence values on different machine learning models, such as DNN, logistic regressions, etc.
Despite the attacks exploiting the privacy leakage in the training sets, Hua et al~\cite{HuaZS18} presented a novel attack to reverse engineer the underlying network information. 
They utilize the memory accessing patterns to infer the network structures, such as number of layers, the feature map sizes of each layer. They also showed the values of weights can be recovered if the memory access can be observed during a ``zero pruning'' stage.
The main difference of our attack from above-mentioned ones is the attack target. Our proposed attack try to explore the leakage at inference stage. Instead of training set samples or network models, we can recover an image for runtime inputs using the power side channel.
\noindent \textbf{Power Side-channel Attack: }
The power side-channel leakage can be exploited to recover the secret keys in cryptographic devices. 
By analyzing the difference of multiple power traces with diverse inputs, attackers are able to uncover the secret key in widely used symmetric encryption standards, such as DES~\cite{KocherJJ99} and AES~\cite{BrierCO04, gierlichs2008mutual,Messerges00}.
Eisenbarth~\cite{EisenbarthPW10} and Msgna~\cite{MsgnaMM13} showed they can recover the instruction type executed by processor via power side channel using hidden Markov model. 
Liu et al~\cite{LiuWZZXX16} managed to accurately locate each instruction instance during execution with a modified Viterbi algorithm. Building a ``power template'' is a common way used to break secret keys in cryptographic systems~\cite{RechbergerO04, choudary2013efficient}. Similar to our proposed attack, they firstly estimate a leakage model from the collected power traces and the secret keys and then at runtime, using the leakage model they predict the keys from the online traces. Though the general procedure is similar, the difficulty in building ``power traces'' is to find a proper attacking surface. In template attacks, they need to identify power traces which only correlate with a limited number of the key bits. In our attack, we found the convolutional unit as the appropriate attack target and raised the attacking method accordingly.


\section{Conclusion}\label{sec:conclusion}
In this paper, we demonstrate the first power side channel attack on an FPGA-based convolutional neural network accelerator. Its input image is successfully recovered using the power traces measured for inferencing operation. 
In the attack, we firstly filter out the noises and distortions in power measurement process.
Then we consider two attacking scenarios for adversaries of different abilities. They can either passively eavasdrop the power side channel or additionally profiling the correlation between power signals and image pixels.
For the two adversaries, we propose two methods respectively: 
background detection and power template, to recover the input image in different granularity.
We demonstrate the practicality of our proposed attack on an accelerator executing classification task for hand-written digits in MNIST datasets and the experimental results show we achieve high recognition accuracy.


\section*{Acknowledgement}\label{sec:ack}
This work was supported in part by the General Research Fund (GRF) of Hong Kong Research Grants Council (RGC) under Grant No. 14205018 and in part by National Natural Science Foundation of China under Grant No. 61432017 and No. 61532017.

\appendix
\section{Preliminaries}\label{sec:appendix-background}
In this section, we first review the concept of convolutional neural network (CNN), and then introduce the architecture of typical CNN accelerators 
and finally discuss the basics on power side-channel leakage.
\subsection{Convolutional Neural Network}
Convolutional Neural Network (CNN)~\cite{lecun1995convolutional} is a neural network architecture used for image applications. 
It is constructed by a pipeline of sequentially connected layers and may consist of four types of computation: convolution, pooling, normalization and full-connected. The structure of the network, such as total number of layers and the type of computation in each layer, is determined by designers prior to the training stage. Then the parameters in each layer, namely weights, are acquired through dedicated training algorithms.
In the inference stage with structure and weights ascertained,
CNN can make predictions with the input images.
In particular, the input of \emph{first layer} of CNN is \emph{image itself} and the computation in the first layer is usually \emph{convolution}. 

\begin{figure*}[htpb]
  \centering
  \includegraphics[width=1.0\linewidth]{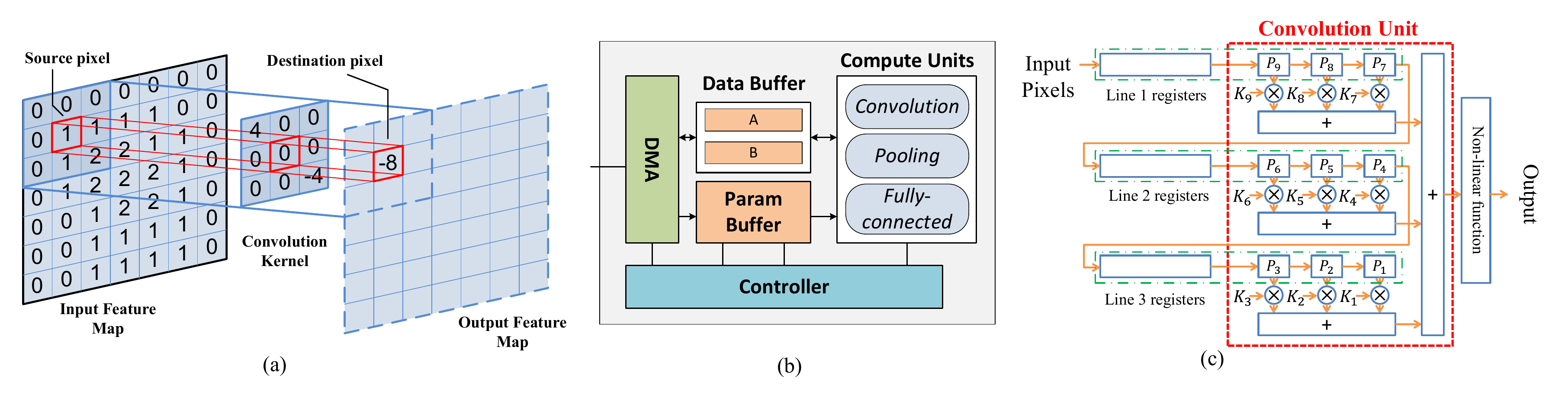}
  \caption{(a) 2D Convolution operation in CNN. (b) Architecture of CNN implementation in ~\cite{ZhaoSZXLSGZ17}. (c) Structure of line buffer}
  \label{fig:bnn_fpga}
\end{figure*}

As our focus is on the convolution layer of CNN, here we briefly introduce its details and illustrate the calculation in Fig.~\ref{fig:bnn_fpga} (a). The input to the convolution layer is an image of size $X \times Y \times M$ and we call the $X \times Y$ 2D pixel array \emph{feature map}. 
For each input feature map, to calculate the pixel value of the output feature map, a kernel (or filter) of size $K_x \times K_y$ is applied to construct a convolutional window for each input pixel capturing its neighbors. We then get an output feature map with the convolutional window sliding by steps of $S_x$ and $S_y$ in two directions of the input feature map. 
We can represent the convolution operation formally with following formula:
\begin{equation}
  O_{x, y}^{j} = f(\sum_{i=1}^{M}(\beta^{i, j} + \sum_{a=0}^{K_x - 1}\sum_{b=0}^{K_y-1}\omega^{i, j}_{a, b} \times I_{xS_x + a, yS_y + b}^{i})),
\end{equation}
where the $O_{x, y}^j$ is the pixel value of position $(x, y)$ in $j$-th output feature map, $\omega^{i,j}$ and $\beta^{i, j}$ are the kernel and bias value between the $i$-th input feature map and the $j$-th output feature map respectively, and $f(\cdot)$ is a non-linear activation function such as $tanh$ or $sigmoid$.


\subsection{CNN Accelerator Design}\label{subsec:cnn_fpga}
An accelerator is usually used in the inference stage to boost the computational efficiency in a number of low-power platforms. 
The accelerators are usually implemented by dedicated hardware, such as FPGA and ASIC and there are a number of designs available~\cite{qiu2016going, conti2015ultra, zhang2017frequency} in both academia and industry. The general architecture of these accelerators is similar, as shown in Fig.~\ref{fig:bnn_fpga} (b) wherein typically five components are involved: Direct Memory Access (DMA), controller, data buffers, parameter buffers and compute units. DMA is used for the data transmission with main processors while controller is responsible for coordinating computation tasks among components. The parameter buffers store the weights used in the CNN model and shall be ready prior to any inference operation.
The data buffer stores the input feature maps for every layer and caches the output feature map from computing units to be used in the next layer.
Compute units contain dedicated hardware to accelerate different operations in the neural network, e.g., convolution, pooling, etc.

Specifically, as the target of our attack is the convolutional layer in the CNN, we present the detailed design for the convolution operation in the compute unit. \emph{Line buffer}~\cite{BosiBS99} is an efficient hardware structure to implement convolution and it has been adopted by a number of CNN accelerators~\cite{qiu2016going, conti2015ultra, zhang2017frequency}.
Fig.~\ref{fig:bnn_fpga} (c) shows the structure of line buffer to execute 2D convolution with a $3 \times 3$ kernel. There are three line registers to compute the convolution with a kernel of size $3 \times 3$ as we need to cache the pixel values in recent three rows of the image. The length of each line is equivalent to the row size of input image.
The convolution is achieved by a set of dedicated hardware multiplier and adders.
At each cycle, one pixel is put into the line buffer, and a $3 \times 3$ convolution is computed. The intermediate result is passed through a non-linear function to generate one output value per cycle.
If the input image contains several channels (e.g., three channels for an RGB image), multiple instances of line buffer are synthesized for parallel procession.
When all input pixels are processed by the line buffer, one output feature map is finished and stored on the data buffer for processing in the next round.
The above-mentioned procedure is repeated until we generate all output feature maps with different kernels.
As we can see from the operation of line buffer, at each cycle, the output only depends on a limited number of input pixels (inside the convolution window), which serves as the foundation to efficiently launch our proposed attack.

In this paper, we follow the implementation proposed by Zhao et al~\cite{ZhaoSZXLSGZ17} who implement accelerator for a compressed version of CNN~\cite{HubaraCSEB16} on FPGA. The convolution unit in their proposed architecture is based on the line buffer.
In their neural accelerator, the parameters and activations inside the network model is limited to either 1 or -1 so that the weights of compressed network can be completely stored inside the RAM of FPGA.

\subsection{Basics on Power Side Channel}\label{subsec:basic_power}
\noindent
\textbf{Power Constitution and Measurement:}
The power consumption of circuits can be divided into two categories: static and dynamic.
Static power consumption arises from the leakage current of transistors and is typically very low. Dynamic power consumption comes from internal transitions of transistors which closely relate to its input data and it usually dominates the total power consumption in its magnitude. To measure the power consumption, a 1$\Omega$ resistor is placed on the power supply line and the voltage drop on it is measured using a high-resolution oscilloscope.


\begin{table}[htbp]
\centering
\caption{Estimated power consumption for line buffer}
\label{tbl:line_power}
\begin{threeparttable}
\begin{tabular}{|l|c|c|}
\hline
\multicolumn{1}{|c|}{\begin{tabular}[c]{@{}c@{}}Line Buffer \\ Configuration\end{tabular}} & \begin{tabular}[c]{@{}c@{}}Convolution Unit \\ Power Estimation\end{tabular} & \begin{tabular}[c]{@{}c@{}}Total Power \\ Estimation\end{tabular} \\
\hline
IC: 1, LS: 28, KS: 3 x 3 & 0.57mW & 0.67mW \\
\hline
IC: 1, LS: 32, KS: 3 x 3 & 0.64mW & 0.79mW \\
\hline
IC: 1, LS: 28, KS: 5 x 5 & 1.25mW & 1.51mW \\
\hline
IC: 3, LS: 28, KS: 3 x 3 & 1.78mW & 2.07mW \\
\hline
\end{tabular}
\begin{tablenotes}
\item[*] IC -- Image Channel, LS -- Line Size, KS -- Kernel Size
\end{tablenotes}
\end{threeparttable}
\end{table}

\noindent
\textbf{Power Consumption of Line Buffer: }
As the line buffer is the main attack target, we estimated the power consumption of the convolution unit and total power consumption with Xilinx XPower Analyzer, a software power emulator for FPGA. We implemented the line buffer in RTL with various configurations and the result is shown in Table~\ref{tbl:line_power}, wherein the convolution unit dominates the total power consumption.
To be specific, we implemented four common configurations of line buffer: three of them have only one input channel, but the line size is 28 and 32 respectively and kernel size can be either 3x3 or 5x5. The last configuration is of three input channels, its line size and filter size are identical to that in the first row. From the statistics of Table~\ref{tbl:line_power}, the power consumption of the convolutional unit increases significantly due to the increase of kernel size and input channel. It is because in these two cases, the pixels involved in the convolution unit increases. The change of line size does not affect much of the power consumed by convolution units.
Whatever the configuration, the power in the convolution unit occupies more than 80\% of the total consumption. Therefore, we can regard the measured power as a coarse-grain estimate for the power of convolution unit.


\section{Discussion and Future Work}\label{sec:discussion}
In this section, we first discuss the applicability of our proposed attack and the attack target of background detection method. 
We also discuss the countermeasures, limitations and future work.

\noindent \textbf{Applicability: }
Though we evaluate our power side-channel attack on the accelerator implemented on FPGA, the actual attack target is the structure of \emph{line buffer} where we exploit the power consumption with the sliding convolutional window over the input image. 
Thus our attack is applicable for whatever designs adopting the line buffer to execute the convolutional operation.
Though line buffer is not suitable for DNN system on CPU or GPU, it enjoys popularity among a variety of FPGA- or ASIC-based neural network accelerators~\cite{qiu2016going, conti2015ultra, zhang2017frequency}. 
Considering the promising future application of neural accelerators, the proposed attack is a severe threat for the security of them.

\noindent \textbf{Attack Target of Background Detection: }
Firstly, the background detection method proposed in Section~\ref{sec:background_detection} is not guaranteed to find all pixels in pure background because its recovery granularity is limited by the kernel size. Thus, the background detection method can fail to recover the images with a messy background. Secondly, the threshold used in background detection is determined by the sharp descend of cycle counts of power consumed per cycle. We may not be able to observe the decline if the number of background pixels is far less than the foreground pixels. To summarize, the background detection method can recover the images which contains a pure and relatively large background region.

\begin{figure}
    \centering
    \includegraphics[width=0.8\linewidth]{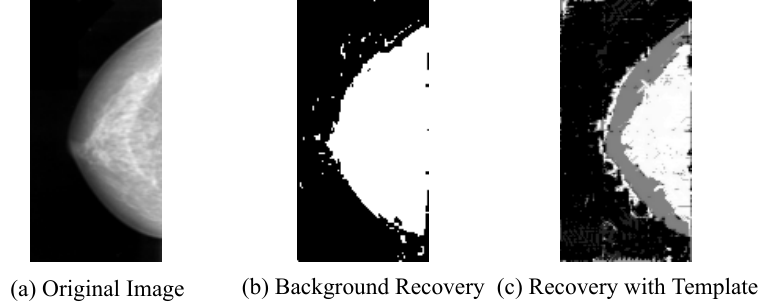}
    \caption{Results on a Mammographic image}
    \label{fig:case_extension}
\end{figure}

\noindent \textbf{Extension to other datasets: }
In our proposed attack, we built our template with MNIST dataset and verify the runtime results with it. It is essential to discuss whether our attack method is still applicable to other datasets. Ideally, the recovery template is built from data inside the convolutional window and the corresponding power consumption, which is independent from the chosen dataset if the template is built in a way avoids overfitting. To demonstrate the effectiveness of our proposed attack, we launch both background detection and power template attacks on an image extracted from the Digital Database for Screening Mammography (DDSM)~\cite{HeathBKKMCM98}, a medical image dataset for mammography research. Fig.~\ref{fig:case_extension} (a) shows the original image (resized to 168 $\times$ 84) from the dataset. It is a side-view radiology image of human breast. Fig.~\ref{fig:case_extension} (b) and Fig.~\ref{fig:case_extension} (c) illustrate the recovered image using background detection and power template respectively. The template used for this demonstration is the same with that we used to recover the MNIST images in Section~\ref{subsec:pixel_attack}. From the recovered image, we can see the shape of breast is pretty much preserved in Fig.~\ref{fig:case_extension} (b), thus for adversary the shape can be used to infer the potential scanned organ of the patient. This leakage somehow reveals the privacy of the patient. We can also get more details in the image recovered from power template. The pixels gradually get lighter from the border part of the breast to the center part, which is consistent with the original image. Though the quality of recovered image cannot reflect every detail of the original image, it still can be used to deduce private information by adversaries.

\noindent \textbf{Countermeasures: }
The most straight forward way to prevent the side channel attacker is to add noises on power side channel, but it does not grant strong guarantees on the privacy protection as noises can still be somehow cancelled with its distribution. For performance and security reason, countermeasures against power side channel attack can be implemented by mainly two ways: \emph{random masking} and \emph{random scheduling}.
Random masking breaks the correlation between the power consumption and the sensitive data by masking the intermediate result with a random number. For instance, before the convolution, each pixel value used for computation is added with a random number. Then after the convolution result is obtained, the result is subtracted by the sum of these random numbers weighted by the convolution kernel.
Random scheduling is effective against active attackers who utilize power from multiple kernels. If the convolution computation for each kernel is executed in a random order rather than sequentially, active adversaries will not be able to build an accurate power feature vector and they can fail in producing a recognizable image.

\noindent \textbf{Limitations and Future Work: }
As mentioned in above subsections, the current proposed attacking algorithm only works for designs based on the line buffer (e.g., on FPGA or ASIC) and it cannot be directly applicable for GPU or TPUs. The attacks can also be defended by simple randomization techniques as discussed in the previous subsection. 
Thirdly, our proposed attack algorithm is currently profiling the power consumption with images coming from the same sampling set. They inherently resemble each other so that we can achieve high recognition accuracy with relatively low overhead. On the other hand, the images adopted in the experiment are either quasi-binary (MNIST) or simple gray-scale images (mammographic pictures). If the attackers target at images with multiple input channel (e.g., color images) or they are not able to get the input images with same distribution of attack target in the profiling phase, more data need to be enrolled to achieve acceptable results. Thus, it is essential to handle the performance problem incurred by complex image recovery task and limited capability of obtaining data from similar distribution.
We may resort to following techniques to tackle the problem: we can use PCA to compress the power feature vector and related pixel values to reduce the size of the pre-built power template and use SVM or random forest to choose candidates in actual attacks. We plan to incorporate them in our future work and validate our method on more complicated datasets, such as CIFAR-10 or even ImageNet.

\section{Attack results on the MNIST dataset}\label{sec:appendix-mnist}
\begin{figure*}
  \centering
  \includegraphics[width=1.0\textwidth]{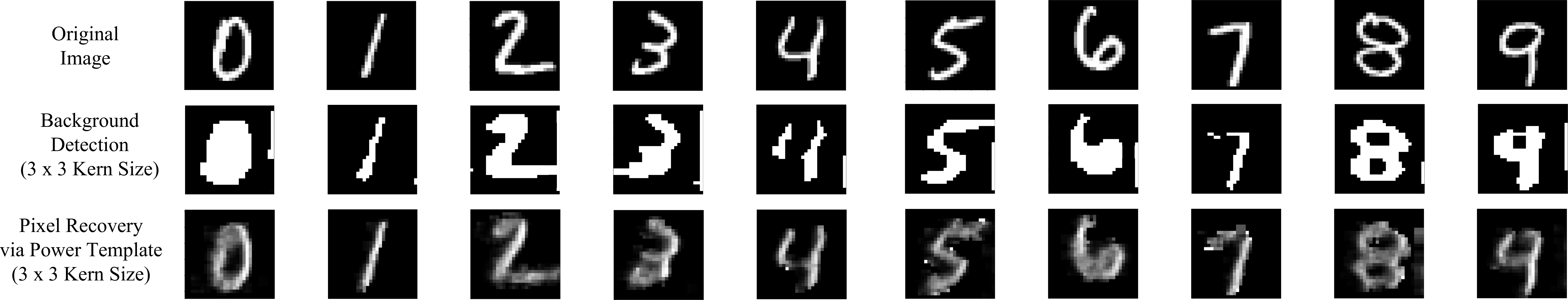}
  \caption{Recovered image of correctly classified digits.}
  \label{fig:mnist_all_digits}
\end{figure*}

The recovered image of from power side channel is illustrated in Fig.~\ref{fig:mnist_all_digits}. For two recovery methods, we select the correctly classified images with the same input image so that we can compare the quality of recovered images directly.


\begin{thebibliography}{39}

  
  \ifx \showCODEN    \undefined \def \showCODEN     #1{\unskip}     \fi
  \ifx \showDOI      \undefined \def \showDOI       #1{#1}\fi
  \ifx \showISBNx    \undefined \def \showISBNx     #1{\unskip}     \fi
  \ifx \showISBNxiii \undefined \def \showISBNxiii  #1{\unskip}     \fi
  \ifx \showISSN     \undefined \def \showISSN      #1{\unskip}     \fi
  \ifx \showLCCN     \undefined \def \showLCCN      #1{\unskip}     \fi
  \ifx \shownote     \undefined \def \shownote      #1{#1}          \fi
  \ifx \showarticletitle \undefined \def \showarticletitle #1{#1}   \fi
  \ifx \showURL      \undefined \def \showURL       {\relax}        \fi
  \providecommand\bibfield[2]{#2}
  \providecommand\bibinfo[2]{#2}
  \providecommand\natexlab[1]{#1}
  \providecommand\showeprint[2][]{arXiv:#2}
  
  \bibitem[\protect\citeauthoryear{Abadi, Chu, Goodfellow, McMahan, Mironov,
    Talwar, and Zhang}{Abadi et~al\mbox{.}}{2016}]%
          {AbadiCGMMT016}
  \bibfield{author}{\bibinfo{person}{Mart{\'{\i}}n Abadi}, \bibinfo{person}{Andy
    Chu}, \bibinfo{person}{Ian~J. Goodfellow}, \bibinfo{person}{H.~Brendan
    McMahan}, \bibinfo{person}{Ilya Mironov}, \bibinfo{person}{Kunal Talwar},
    {and} \bibinfo{person}{Li Zhang}.} \bibinfo{year}{2016}\natexlab{}.
  \newblock \showarticletitle{Deep Learning with Differential Privacy}. In
    \bibinfo{booktitle}{\emph{Proc. of {ACM} {SIGSAC} Conference on Computer and
    Communications Security ({CCS})}}. \bibinfo{pages}{308--318}.
  \newblock
  
  
  \bibitem[\protect\citeauthoryear{Bosi, Bois, and Savaria}{Bosi
    et~al\mbox{.}}{1999}]%
          {BosiBS99}
  \bibfield{author}{\bibinfo{person}{B. Bosi}, \bibinfo{person}{Guy Bois}, {and}
    \bibinfo{person}{Yvon Savaria}.} \bibinfo{year}{1999}\natexlab{}.
  \newblock \showarticletitle{Reconfigurable pipelined 2-D convolvers for fast
    digital signal processing}.
  \newblock \bibinfo{journal}{\emph{{IEEE} Transactions on {VLSI} Systems}}
    \bibinfo{volume}{7}, \bibinfo{number}{3} (\bibinfo{year}{1999}),
    \bibinfo{pages}{299--308}.
  \newblock
  
  
  \bibitem[\protect\citeauthoryear{Brier, Clavier, and Olivier}{Brier
    et~al\mbox{.}}{2004}]%
          {BrierCO04}
  \bibfield{author}{\bibinfo{person}{Eric Brier}, \bibinfo{person}{Christophe
    Clavier}, {and} \bibinfo{person}{Francis Olivier}.}
    \bibinfo{year}{2004}\natexlab{}.
  \newblock \showarticletitle{Correlation Power Analysis with a Leakage Model}.
    In \bibinfo{booktitle}{\emph{Proc. of Cryptographic Hardware and Embedded
    Systems - {CHES}}}. \bibinfo{pages}{16--29}.
  \newblock
  
  
  \bibitem[\protect\citeauthoryear{Choudary and Kuhn}{Choudary and Kuhn}{2013}]%
          {choudary2013efficient}
  \bibfield{author}{\bibinfo{person}{Omar Choudary} {and}
    \bibinfo{person}{Markus~G Kuhn}.} \bibinfo{year}{2013}\natexlab{}.
  \newblock \showarticletitle{Efficient template attacks}. In
    \bibinfo{booktitle}{\emph{International Conference on Smart Card Research and
    Advanced Applications}}. Springer, \bibinfo{pages}{253--270}.
  \newblock
  
  
  \bibitem[\protect\citeauthoryear{Conti and Benini}{Conti and Benini}{2015}]%
          {conti2015ultra}
  \bibfield{author}{\bibinfo{person}{Francesco Conti} {and} \bibinfo{person}{Luca
    Benini}.} \bibinfo{year}{2015}\natexlab{}.
  \newblock \showarticletitle{A ultra-low-energy convolution engine for fast
    brain-inspired vision in multicore clusters}. In
    \bibinfo{booktitle}{\emph{Design, Automation \& Test in Europe Conference \&
    Exhibition (DATE), 2015}}. IEEE, \bibinfo{pages}{683--688}.
  \newblock
  
  
  \bibitem[\protect\citeauthoryear{Eisenbarth, Paar, and Weghenkel}{Eisenbarth
    et~al\mbox{.}}{2010}]%
          {EisenbarthPW10}
  \bibfield{author}{\bibinfo{person}{Thomas Eisenbarth},
    \bibinfo{person}{Christof Paar}, {and} \bibinfo{person}{Bj{\"{o}}rn
    Weghenkel}.} \bibinfo{year}{2010}\natexlab{}.
  \newblock \showarticletitle{Building a Side Channel Based Disassembler}.
  \newblock \bibinfo{journal}{\emph{Trans. Computational Science}}
    \bibinfo{volume}{10} (\bibinfo{year}{2010}), \bibinfo{pages}{78--99}.
  \newblock
  
  
  \bibitem[\protect\citeauthoryear{Fredrikson, Jha, and Ristenpart}{Fredrikson
    et~al\mbox{.}}{2015}]%
          {FredriksonJR15}
  \bibfield{author}{\bibinfo{person}{Matt Fredrikson}, \bibinfo{person}{Somesh
    Jha}, {and} \bibinfo{person}{Thomas Ristenpart}.}
    \bibinfo{year}{2015}\natexlab{}.
  \newblock \showarticletitle{Model Inversion Attacks that Exploit Confidence
    Information and Basic Countermeasures}. In \bibinfo{booktitle}{\emph{Proc. of
    {ACM} {SIGSAC} Conference on Computer and Communications Security {(CCS)}}}.
    \bibinfo{pages}{1322--1333}.
  \newblock
  
  
  \bibitem[\protect\citeauthoryear{Fredrikson, Lantz, Jha, Lin, Page, and
    Ristenpart}{Fredrikson et~al\mbox{.}}{2014}]%
          {FredriksonLJLPR14}
  \bibfield{author}{\bibinfo{person}{Matthew Fredrikson}, \bibinfo{person}{Eric
    Lantz}, \bibinfo{person}{Somesh Jha}, \bibinfo{person}{Simon Lin},
    \bibinfo{person}{David Page}, {and} \bibinfo{person}{Thomas Ristenpart}.}
    \bibinfo{year}{2014}\natexlab{}.
  \newblock \showarticletitle{Privacy in Pharmacogenetics: An End-to-End Case
    Study of Personalized Warfarin Dosing}. In
    \bibinfo{booktitle}{\emph{Proceedings of the 23rd {USENIX} Security
    Symposium, San Diego, CA, USA, August 20-22, 2014.}} \bibinfo{pages}{17--32}.
  \newblock
  
  
  \bibitem[\protect\citeauthoryear{Gierlichs, Batina, Tuyls, and
    Preneel}{Gierlichs et~al\mbox{.}}{2008}]%
          {gierlichs2008mutual}
  \bibfield{author}{\bibinfo{person}{Benedikt Gierlichs}, \bibinfo{person}{Lejla
    Batina}, \bibinfo{person}{Pim Tuyls}, {and} \bibinfo{person}{Bart Preneel}.}
    \bibinfo{year}{2008}\natexlab{}.
  \newblock \showarticletitle{Mutual information analysis}.
  \newblock \bibinfo{journal}{\emph{Cryptographic Hardware and Embedded
    Systems--CHES 2008}} (\bibinfo{year}{2008}), \bibinfo{pages}{426--442}.
  \newblock
  
  
  \bibitem[\protect\citeauthoryear{Heath, Bowyer, Kopans, Jr., Moore, Chang, and
    Munishkumaran}{Heath et~al\mbox{.}}{1998}]%
          {HeathBKKMCM98}
  \bibfield{author}{\bibinfo{person}{M. Heath}, \bibinfo{person}{K. Bowyer},
    \bibinfo{person}{Daniel~B. Kopans}, \bibinfo{person}{P.~Kegelmeyer Jr.},
    \bibinfo{person}{Richard~H. Moore}, \bibinfo{person}{K. Chang}, {and}
    \bibinfo{person}{S. Munishkumaran}.} \bibinfo{year}{1998}\natexlab{}.
  \newblock \showarticletitle{Current Status of the Digital Database for
    Screening Mammography}. In \bibinfo{booktitle}{\emph{Digital Mammography,
    Fourth International Workshop on Digital Mammograph, {IWDM} 1998, Nijmegen,
    The Netherlands, June 1998}}. \bibinfo{pages}{457--460}.
  \newblock
  \urldef\tempurl%
  \url{https://doi.org/10.1007/978-94-011-5318-8_75}
  \showDOI{\tempurl}
  
  
  \bibitem[\protect\citeauthoryear{Hua, Zhang, and Suh}{Hua
    et~al\mbox{.}}{2018}]%
          {HuaZS18}
  \bibfield{author}{\bibinfo{person}{Weizhe Hua}, \bibinfo{person}{Zhiru Zhang},
    {and} \bibinfo{person}{G.~Edward Suh}.} \bibinfo{year}{2018}\natexlab{}.
  \newblock \showarticletitle{Reverse engineering convolutional neural networks
    through side-channel information leaks}. In
    \bibinfo{booktitle}{\emph{Proceedings of the 55th Annual Design Automation
    Conference, {DAC} 2018, San Francisco, CA, USA, June 24-29, 2018}}.
    \bibinfo{pages}{4:1--4:6}.
  \newblock
  
  
  \bibitem[\protect\citeauthoryear{Hubara, Courbariaux, Soudry, El{-}Yaniv, and
    Bengio}{Hubara et~al\mbox{.}}{2016}]%
          {HubaraCSEB16}
  \bibfield{author}{\bibinfo{person}{Itay Hubara}, \bibinfo{person}{Matthieu
    Courbariaux}, \bibinfo{person}{Daniel Soudry}, \bibinfo{person}{Ran
    El{-}Yaniv}, {and} \bibinfo{person}{Yoshua Bengio}.}
    \bibinfo{year}{2016}\natexlab{}.
  \newblock \showarticletitle{Binarized Neural Networks}. In
    \bibinfo{booktitle}{\emph{Advances in Neural Information Processing Systems
    29: Annual Conference on Neural Information Processing Systems 2016, December
    5-10, 2016, Barcelona, Spain}}. \bibinfo{pages}{4107--4115}.
  \newblock
  
  
  \bibitem[\protect\citeauthoryear{Inc}{Inc}{2017a}]%
          {BIGML}
  \bibfield{author}{\bibinfo{person}{BigML Inc}.}
    \bibinfo{year}{2017}\natexlab{a}.
  \newblock \bibinfo{title}{{BigML}}.
  \newblock
  \newblock
  \urldef\tempurl%
  \url{https://www.bigml.com/}
  \showURL{%
  \tempurl}
  
  
  \bibitem[\protect\citeauthoryear{Inc}{Inc}{2017b}]%
          {MSAZURE}
  \bibfield{author}{\bibinfo{person}{Microsoft Inc}.}
    \bibinfo{year}{2017}\natexlab{b}.
  \newblock \bibinfo{title}{Microsoft Azure Machine Learning}.
  \newblock
  \newblock
  \urldef\tempurl%
  \url{https://azure.microsoft.com/en-us/services/machine-learning/}
  \showURL{%
  \tempurl}
  
  
  \bibitem[\protect\citeauthoryear{Inc}{Inc}{2017c}]%
          {XILINXSP}
  \bibfield{author}{\bibinfo{person}{Xilinx Inc}.}
    \bibinfo{year}{2017}\natexlab{c}.
  \newblock \bibinfo{title}{Xilinx Spartan-6 FPGA family}.
  \newblock
  \newblock
  \urldef\tempurl%
  \url{https://www.xilinx.com/products/silicon-devices/fpga/spartan-6.html}
  \showURL{%
  \tempurl}
  
  
  \bibitem[\protect\citeauthoryear{Instruments}{Instruments}{2017}]%
          {NI}
  \bibfield{author}{\bibinfo{person}{National Instruments}.}
    \bibinfo{year}{2017}\natexlab{}.
  \newblock \bibinfo{title}{NI Multisim}.
  \newblock
  \newblock
  \urldef\tempurl%
  \url{www.ni.com/multisim/}
  \showURL{%
  \tempurl}
  
  
  \bibitem[\protect\citeauthoryear{Kocher, Jaffe, and Jun}{Kocher
    et~al\mbox{.}}{1999}]%
          {KocherJJ99}
  \bibfield{author}{\bibinfo{person}{Paul~C. Kocher}, \bibinfo{person}{Joshua
    Jaffe}, {and} \bibinfo{person}{Benjamin Jun}.}
    \bibinfo{year}{1999}\natexlab{}.
  \newblock \showarticletitle{Differential Power Analysis}. In
    \bibinfo{booktitle}{\emph{Proc. of Annual International Cryptology Conference
    - {CRYPTO} '99}}. \bibinfo{pages}{388--397}.
  \newblock
  
  
  \bibitem[\protect\citeauthoryear{Lab./UEC}{Lab./UEC}{2017a}]%
          {SAKURA17}
  \bibfield{author}{\bibinfo{person}{Satoh Lab./UEC}.}
    \bibinfo{year}{2017}\natexlab{a}.
  \newblock \bibinfo{title}{SAKURA-G}.
  \newblock
  \newblock
  \urldef\tempurl%
  \url{http://satoh.cs.uec.ac.jp/SAKURA/hardware/SAKURA-G.html}
  \showURL{%
  \tempurl}
  
  
  \bibitem[\protect\citeauthoryear{Lab./UEC}{Lab./UEC}{2017b}]%
          {sakuracir}
  \bibfield{author}{\bibinfo{person}{Satoh Lab./UEC}.}
    \bibinfo{year}{2017}\natexlab{b}.
  \newblock \bibinfo{title}{SAKURA: Side-channel AttacK User Reference
    Architecture -- Specification}.
  \newblock
  \newblock
  \urldef\tempurl%
  \url{http://satoh.cs.uec.ac.jp/SAKURA/hardware/SAKURA-G_Spec_Ver1.0_English.pdf}
  \showURL{%
  \tempurl}
  
  
  \bibitem[\protect\citeauthoryear{LeCun, Bengio, et~al\mbox{.}}{LeCun
    et~al\mbox{.}}{1995}]%
          {lecun1995convolutional}
  \bibfield{author}{\bibinfo{person}{Yann LeCun}, \bibinfo{person}{Yoshua
    Bengio}, {et~al\mbox{.}}} \bibinfo{year}{1995}\natexlab{}.
  \newblock \showarticletitle{Convolutional networks for images, speech, and time
    series}.
  \newblock \bibinfo{journal}{\emph{The handbook of brain theory and neural
    networks}} \bibinfo{volume}{3361}, \bibinfo{number}{10}
    (\bibinfo{year}{1995}), \bibinfo{pages}{1995}.
  \newblock
  
  
  \bibitem[\protect\citeauthoryear{LeCun, Cortes, and Burges}{LeCun
    et~al\mbox{.}}{01  }]%
          {MNIST}
  \bibfield{author}{\bibinfo{person}{Yann LeCun}, \bibinfo{person}{Corinna
    Cortes}, {and} \bibinfo{person}{Christopher~J.C. Burges}.}
    \bibinfo{year}{2001--}\natexlab{}.
  \newblock \bibinfo{title}{{THE MNIST DATABASE} of handwritten digits}.
  \newblock \bibinfo{howpublished}{\url{http://yann.lecun.com/exdb/mnist/}}.
  \newblock
  
  
  \bibitem[\protect\citeauthoryear{Liu, Wei, Zhou, Zhang, Xu, and Xu}{Liu
    et~al\mbox{.}}{2016}]%
          {LiuWZZXX16}
  \bibfield{author}{\bibinfo{person}{Yannan Liu}, \bibinfo{person}{Lingxiao Wei},
    \bibinfo{person}{Zhe Zhou}, \bibinfo{person}{Kehuan Zhang},
    \bibinfo{person}{Wenyuan Xu}, {and} \bibinfo{person}{Qiang Xu}.}
    \bibinfo{year}{2016}\natexlab{}.
  \newblock \showarticletitle{On Code Execution Tracking via Power Side-Channel}.
    In \bibinfo{booktitle}{\emph{Proc. of {ACM} {SIGSAC} Conference on Computer
    and Communications Security ({CCS})}}. \bibinfo{pages}{1019--1031}.
  \newblock
  
  
  \bibitem[\protect\citeauthoryear{Messerges}{Messerges}{2000}]%
          {Messerges00}
  \bibfield{author}{\bibinfo{person}{Thomas~S. Messerges}.}
    \bibinfo{year}{2000}\natexlab{}.
  \newblock \showarticletitle{Using Second-Order Power Analysis to Attack {DPA}
    Resistant Software}. In \bibinfo{booktitle}{\emph{Cryptographic Hardware and
    Embedded Systems - {CHES} 2000, Second International Workshop, Worcester, MA,
    USA, August 17-18, 2000, Proceedings}}. \bibinfo{pages}{238--251}.
  \newblock
  
  
  \bibitem[\protect\citeauthoryear{Mohassel and Zhang}{Mohassel and
    Zhang}{2017}]%
          {MohasselZ17}
  \bibfield{author}{\bibinfo{person}{Payman Mohassel} {and}
    \bibinfo{person}{Yupeng Zhang}.} \bibinfo{year}{2017}\natexlab{}.
  \newblock \showarticletitle{SecureML: {A} System for Scalable
    Privacy-Preserving Machine Learning}. In \bibinfo{booktitle}{\emph{Proc. of
    {IEEE} Symposium on Security and Privacy {SP}}}. \bibinfo{pages}{19--38}.
  \newblock
  
  
  \bibitem[\protect\citeauthoryear{Msgna, Markantonakis, and Mayes}{Msgna
    et~al\mbox{.}}{2013}]%
          {MsgnaMM13}
  \bibfield{author}{\bibinfo{person}{Mehari Msgna}, \bibinfo{person}{Konstantinos
    Markantonakis}, {and} \bibinfo{person}{Keith Mayes}.}
    \bibinfo{year}{2013}\natexlab{}.
  \newblock \showarticletitle{The B-Side of Side Channel Leakage: Control Flow
    Security in Embedded Systems}. In \bibinfo{booktitle}{\emph{Security and
    Privacy in Communication Networks - 9th International {ICST} Conference,
    SecureComm 2013, Sydney, NSW, Australia, September 25-28, 2013, Revised
    Selected Papers}}. \bibinfo{pages}{288--304}.
  \newblock
  
  
  \bibitem[\protect\citeauthoryear{O'Haver}{O'Haver}{1997}]%
          {o1997pragmatic}
  \bibfield{author}{\bibinfo{person}{Tom O'Haver}.}
    \bibinfo{year}{1997}\natexlab{}.
  \newblock \bibinfo{title}{A Pragmatic introduction to signal processing}.
  \newblock
  \newblock
  
  
  \bibitem[\protect\citeauthoryear{Papernot, McDaniel, Jha, Fredrikson, Celik,
    and Swami}{Papernot et~al\mbox{.}}{2016a}]%
          {PapernotMJFCS16}
  \bibfield{author}{\bibinfo{person}{Nicolas Papernot},
    \bibinfo{person}{Patrick~D. McDaniel}, \bibinfo{person}{Somesh Jha},
    \bibinfo{person}{Matt Fredrikson}, \bibinfo{person}{Z.~Berkay Celik}, {and}
    \bibinfo{person}{Ananthram Swami}.} \bibinfo{year}{2016}\natexlab{a}.
  \newblock \showarticletitle{The Limitations of Deep Learning in Adversarial
    Settings}. In \bibinfo{booktitle}{\emph{Proc. of {IEEE} European Symposium on
    Security and Privacy, EuroS{\&}P}}. \bibinfo{pages}{372--387}.
  \newblock
  
  
  \bibitem[\protect\citeauthoryear{Papernot, McDaniel, Wu, Jha, and
    Swami}{Papernot et~al\mbox{.}}{2016b}]%
          {PapernotM0JS16}
  \bibfield{author}{\bibinfo{person}{Nicolas Papernot},
    \bibinfo{person}{Patrick~D. McDaniel}, \bibinfo{person}{Xi Wu},
    \bibinfo{person}{Somesh Jha}, {and} \bibinfo{person}{Ananthram Swami}.}
    \bibinfo{year}{2016}\natexlab{b}.
  \newblock \showarticletitle{Distillation as a Defense to Adversarial
    Perturbations Against Deep Neural Networks}. In
    \bibinfo{booktitle}{\emph{Proc. of {IEEE} Symposium on Security and Privacy
    {SP}}}. \bibinfo{pages}{582--597}.
  \newblock
  
  
  \bibitem[\protect\citeauthoryear{Qiu, Wang, Yao, Guo, Li, Zhou, Yu, Tang, Xu,
    Song, et~al\mbox{.}}{Qiu et~al\mbox{.}}{2016}]%
          {qiu2016going}
  \bibfield{author}{\bibinfo{person}{Jiantao Qiu}, \bibinfo{person}{Jie Wang},
    \bibinfo{person}{Song Yao}, \bibinfo{person}{Kaiyuan Guo},
    \bibinfo{person}{Boxun Li}, \bibinfo{person}{Erjin Zhou},
    \bibinfo{person}{Jincheng Yu}, \bibinfo{person}{Tianqi Tang},
    \bibinfo{person}{Ningyi Xu}, \bibinfo{person}{Sen Song}, {et~al\mbox{.}}}
    \bibinfo{year}{2016}\natexlab{}.
  \newblock \showarticletitle{Going deeper with embedded fpga platform for
    convolutional neural network}. In \bibinfo{booktitle}{\emph{Proceedings of
    the 2016 ACM/SIGDA International Symposium on Field-Programmable Gate
    Arrays}}. ACM, \bibinfo{pages}{26--35}.
  \newblock
  
  
  \bibitem[\protect\citeauthoryear{Qualcomm}{Qualcomm}{2017}]%
          {QSSNAP835}
  \bibfield{author}{\bibinfo{person}{Qualcomm}.} \bibinfo{year}{2017}\natexlab{}.
  \newblock \bibinfo{title}{Artificial intelligence tech in Snapdragon 835}.
  \newblock
  \newblock
  \urldef\tempurl%
  \url{https://www.qualcomm.com/snapdragon/artificial-intelligence}
  \showURL{%
  \tempurl}
  
  
  \bibitem[\protect\citeauthoryear{Rafael~Gonzalez and Woods}{Rafael~Gonzalez and
    Woods}{2002}]%
          {rafael2002digital}
  \bibfield{author}{\bibinfo{person}{C Rafael~Gonzalez} {and}
    \bibinfo{person}{Richard Woods}.} \bibinfo{year}{2002}\natexlab{}.
  \newblock \showarticletitle{Digital image processing}.
  \newblock \bibinfo{journal}{\emph{Pearson Education}} (\bibinfo{year}{2002}).
  \newblock
  
  
  \bibitem[\protect\citeauthoryear{Rechberger and Oswald}{Rechberger and
    Oswald}{2004}]%
          {RechbergerO04}
  \bibfield{author}{\bibinfo{person}{Christian Rechberger} {and}
    \bibinfo{person}{Elisabeth Oswald}.} \bibinfo{year}{2004}\natexlab{}.
  \newblock \showarticletitle{Practical Template Attacks}. In
    \bibinfo{booktitle}{\emph{Information Security Applications, 5th
    International Workshop, {WISA} 2004, Jeju Island, Korea, August 23-25, 2004,
    Revised Selected Papers}}. \bibinfo{pages}{440--456}.
  \newblock
  
  
  \bibitem[\protect\citeauthoryear{Shokri, Stronati, Song, and Shmatikov}{Shokri
    et~al\mbox{.}}{2017}]%
          {ShokriSSS17}
  \bibfield{author}{\bibinfo{person}{Reza Shokri}, \bibinfo{person}{Marco
    Stronati}, \bibinfo{person}{Congzheng Song}, {and} \bibinfo{person}{Vitaly
    Shmatikov}.} \bibinfo{year}{2017}\natexlab{}.
  \newblock \showarticletitle{Membership Inference Attacks Against Machine
    Learning Models}. In \bibinfo{booktitle}{\emph{Proc. of {IEEE} Symposium on
    Security and Privacy, {SP}}}. \bibinfo{pages}{3--18}.
  \newblock
  

  \bibitem[\protect\citeauthoryear{Suh, Clarke, Gassend, van Dijk, and
    Devadas}{Suh et~al\mbox{.}}{2003}]%
          {SuhCGDD03}
  \bibfield{author}{\bibinfo{person}{G.~Edward Suh}, \bibinfo{person}{Dwaine~E.
    Clarke}, \bibinfo{person}{Blaise Gassend}, \bibinfo{person}{Marten van Dijk},
    {and} \bibinfo{person}{Srinivas Devadas}.} \bibinfo{year}{2003}\natexlab{}.
  \newblock \showarticletitle{{AEGIS:} architecture for tamper-evident and
    tamper-resistant processing}. In \bibinfo{booktitle}{\emph{Proceedings of the
    17th Annual International Conference on Supercomputing, {ICS} 2003, San
    Francisco, CA, USA, June 23-26, 2003}}. \bibinfo{pages}{160--171}.
  \newblock
  
  \newpage
  
  \bibitem[\protect\citeauthoryear{Sze, Chen, Yang, and Emer}{Sze
    et~al\mbox{.}}{2017}]%
          {sze2017efficient}
  \bibfield{author}{\bibinfo{person}{Vivienne Sze}, \bibinfo{person}{Yu-Hsin
    Chen}, \bibinfo{person}{Tien-Ju Yang}, {and} \bibinfo{person}{Joel Emer}.}
    \bibinfo{year}{2017}\natexlab{}.
  \newblock \showarticletitle{Efficient processing of deep neural networks: A
    tutorial and survey}.
  \newblock \bibinfo{journal}{\emph{arXiv preprint arXiv:1703.09039}}
    (\bibinfo{year}{2017}).
  \newblock
  
  
  \bibitem[\protect\citeauthoryear{Tektronics}{Tektronics}{2017}]%
          {TEKMDO}
  \bibfield{author}{\bibinfo{person}{Tektronics}.}
    \bibinfo{year}{2017}\natexlab{}.
  \newblock \bibinfo{title}{MDO3000 Mixed Domain Oscilloscope}.
  \newblock
  \newblock
  \urldef\tempurl%
  \url{http://www.tek.com/oscilloscope/mdo3000-mixed-domain-oscilloscope}
  \showURL{%
  \tempurl}
  
  
  \bibitem[\protect\citeauthoryear{Tram{\`{e}}r, Zhang, Juels, Reiter, and
    Ristenpart}{Tram{\`{e}}r et~al\mbox{.}}{2016}]%
          {TramerZJRR16}
  \bibfield{author}{\bibinfo{person}{Florian Tram{\`{e}}r}, \bibinfo{person}{Fan
    Zhang}, \bibinfo{person}{Ari Juels}, \bibinfo{person}{Michael~K. Reiter},
    {and} \bibinfo{person}{Thomas Ristenpart}.} \bibinfo{year}{2016}\natexlab{}.
  \newblock \showarticletitle{Stealing Machine Learning Models via Prediction
    APIs}. In \bibinfo{booktitle}{\emph{25th {USENIX} Security Symposium,
    {USENIX} Security 16, Austin, TX, USA, August 10-12, 2016.}}
    \bibinfo{pages}{601--618}.
  \newblock
  
  
  \bibitem[\protect\citeauthoryear{Zhang and Prasanna}{Zhang and
    Prasanna}{2017}]%
          {zhang2017frequency}
  \bibfield{author}{\bibinfo{person}{Chi Zhang} {and} \bibinfo{person}{Viktor
    Prasanna}.} \bibinfo{year}{2017}\natexlab{}.
  \newblock \showarticletitle{Frequency domain acceleration of convolutional
    neural networks on CPU-FPGA shared memory system}. In
    \bibinfo{booktitle}{\emph{Proceedings of the 2017 ACM/SIGDA International
    Symposium on Field-Programmable Gate Arrays}}. ACM, \bibinfo{pages}{35--44}.
  \newblock
  
  
  \bibitem[\protect\citeauthoryear{Zhao, Song, Zhang, Xing, Lin, Srivastava,
    Gupta, and Zhang}{Zhao et~al\mbox{.}}{2017}]%
          {ZhaoSZXLSGZ17}
  \bibfield{author}{\bibinfo{person}{Ritchie Zhao}, \bibinfo{person}{Weinan
    Song}, \bibinfo{person}{Wentao Zhang}, \bibinfo{person}{Tianwei Xing},
    \bibinfo{person}{Jeng{-}Hau Lin}, \bibinfo{person}{Mani~B. Srivastava},
    \bibinfo{person}{Rajesh Gupta}, {and} \bibinfo{person}{Zhiru Zhang}.}
    \bibinfo{year}{2017}\natexlab{}.
  \newblock \showarticletitle{Accelerating Binarized Convolutional Neural
    Networks with Software-Programmable FPGAs}. In
    \bibinfo{booktitle}{\emph{Proc. of the {ACM/SIGDA} International Symposium on
    Field-Programmable Gate Arrays {FPGA}}}. \bibinfo{pages}{15--24}.
  \newblock
  
  
  \end{thebibliography}


\end{document}